\documentclass[journal,transmag]{IEEEtran}
\usepackage{cite}
\usepackage{mathtools}
\ifCLASSINFOpdf
  \usepackage[pdftex]{graphicx}
\else
\fi
\usepackage{amsmath}

\usepackage{amssymb}
\usepackage{units}
\ifCLASSOPTIONcompsoc
 \usepackage[caption=false,font=normalsize,labelfont=sf,textfont=sf]{subfig}
\else
 \usepackage[caption=false,font=footnotesize]{subfig}
\usepackage{url}
\begin{document}
\title{Non-Contact Manipulation of Induced Magnetic Dipoles}


\author{\IEEEauthorblockN{Seth Stewart\IEEEauthorrefmark{1},
Joseph Pawelski\IEEEauthorrefmark{2},
Steve Ward\IEEEauthorrefmark{2}, and 
Andrew J. Petruska\IEEEauthorrefmark{1},~\IEEEmembership{Senior~Member,~IEEE}}
\IEEEauthorblockA{\IEEEauthorrefmark{1} Colorado School of Mines, Golden, CO 80401}
\IEEEauthorblockA{\IEEEauthorrefmark{2} CisLunar Industries, Loveland, CO 80537}
\thanks{Manuscript received XXXX XX, 202X; revised XXXX XX, 202X
Corresponding author: S. Stewart (email: sstewart@mines.edu).}}

\markboth{Journal of \LaTeX\ Class Files,~Vol.~XX, No.~X, XX~202X}%
{Shell \MakeLowercase{\textit{et al.}}: Bare Demo of IEEEtran.cls for IEEE Transactions on Magnetics Journals}



% for Transactions on Magnetics papers, we must declare the abstract and
% index terms PRIOR to the title within the \IEEEtitleabstractindextext
% IEEEtran command as these need to go into the title area created by
\IEEEtitleabstractindextext{%
\begin{abstract}
Extending the field of magnetic manipulation to conductive, non-magnetic objects opens the door for a wide array of applications previously limited to hard or soft magnetic materials. Of particular interest is the recycling of space debris through the use of oscillating magnetic fields, which represent a cache of raw materials in an environment particularly suited to the low forces generated from inductive magnetic manipulation. Building upon previous work that demonstrated 3D open-loop position control by leveraging the opposing dipole moment created from induced eddy currents, this work demonstrates closed-loop position control of a semi-buoyant aluminum sphere in lab tests, and the efficacy of varying methods for force inversion is explored. The closed-loop methods represent a critical first step towards wider applications for 3-DOF position control of induced magnetic dipoles.
  
\end{abstract}

}

\maketitle

\IEEEdisplaynontitleabstractindextext

% \fi
\IEEEpeerreviewmaketitle

\section{Introduction} \label{sec:Introduction}

\IEEEPARstart{T}{he} previous decade has seen a significant reduction in the cost of launching materials into space, and this new reality has been accompanied by an increase in man-made satellites occupying low Earth orbit \cite{jones2018recent}\cite{villela2019towards}. This renewed interest in scientific endeavors in space also brings about concerns of the formation of a debris belt that would jeopardize humanity's future in space \cite{kessler1978collision}\cite{kessler2010kessler}. Recognizing the wealth of materials that defunct satellites and rocket bodies represent, some have suggested that recycling this space debris could help to alleviate the problems of space junk and develop a space economy in one fell swoop \cite{schroeder2023space}. 

 To that end, several methods of capturing and manipulating debris in microgravity have been suggested \cite{jiang2017robotic}\cite{mark2019review}, with an the electromagnetic field being particularly attractive due to its ability to act as both a source of positioning and heating \cite{stewart2025open}. Because of the inherently non-contact nature of magnetic manipulation, it has seen much of its application in the field of medicine \cite{nelson2010microrobots, keller2012method, hong2020magnetic}. This ability to gently maneuver objects from a distance makes it also suitable for manipulation of molten metals in microgravity, where decanting into a mold would be impractical. 

  %%%%%%%%%%%%%%%%%%%%%%%%%%%%%%%%%%%%%%%%%%%%%%%%%%%
% %%%%%%%%%%%%%%%%%%%%%%%%%%%%%%%%%%%%%%%%%%%%%%%%%%%
\begin{figure}[t!]
    \centering
    \includegraphics[width=9cm]{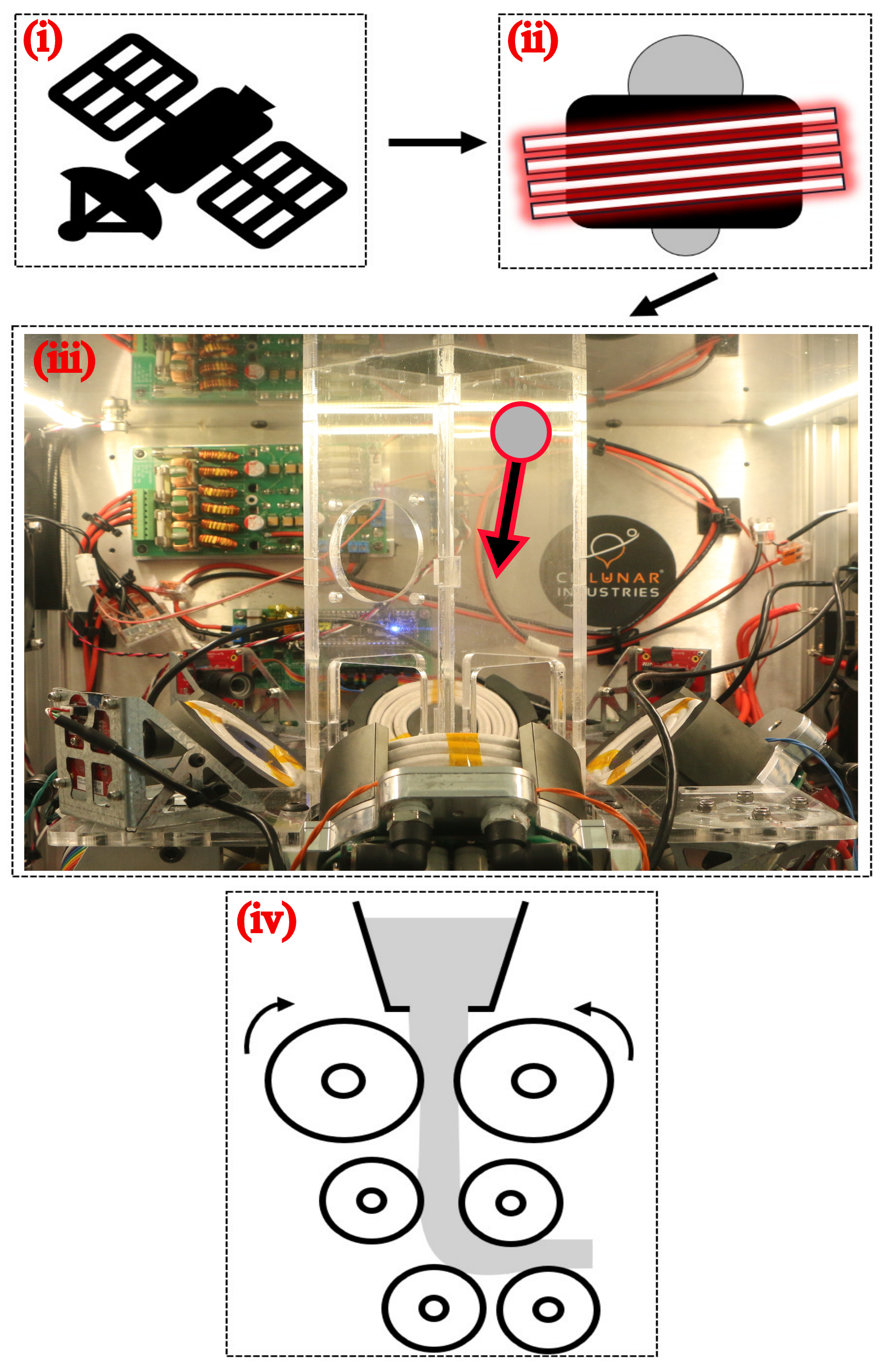}
    \caption{The four-stage recycling process as imagined by this research group. [i] Space debris is captured in-orbit and stripped for usable raw materials. [ii] Conductive materials (mainly aluminum) are melted in a furnace, here imagined as an induction heater. [iii] Molten aluminum droplets are carried from the furnace and positioned using time-varying magnetic fields that induce eddy currents in the molten sample, allowing for position control and thermal management as the metal moves from furnace to site of casting. [iv] Molten samples arrive at the site of casting, which re-purposes the raw materials into a usable material in-orbit. Here, casting is imagined as a continuous-casting process. } 
    \label{fig:recyclingProcess}

\end{figure}
% %%%%%%%%%%%%%%%%%%%%%%%%%%%%%%%%%%%%%%%%%%%%%%%%%%%
% %%%%%%%%%%%%%%%%%%%%%%%%%%%%%%%%%%%%%%%%%%%%%%%%%%%

 Recycling space debris can be seen as a four-step process, as outlined in Figure \ref{fig:recyclingProcess}, where space debris is first collected and sorted into usable raw material. This material is placed in an induction furnace to melt and then is guided using the methods described in this paper from a furnace to a casting process. Figure \ref{fig:recyclingProcess} (iv) represents the casting process as a continuous-cast where large rollers act as a site for removing heat and forming the metal \cite{louhenkilpi2024continuous}, though it can be easily extended to the use of static-molds or even an additive manufacturing procedure. This steering from the furnace to cast (step (iii) in Figure \ref{fig:recyclingProcess}) is a critical step in the process, as dexterous-manipulation of molten metals in the absence of gravity is yet an unsolved problem, and could represent a step in the process where components are at risk of being damaged by molten metals. Further, if not properly controlled, these metal scraps are at risk of being jettisoned from the process and once again becoming uncontrolled space debris. As an added benefit of non-contact manipulation, contamination from with contacting a crucible is largely avoided \cite{okress1952electromagnetic}, allowing for enhanced control of nucleation sites and allowing the creation of exotic metals only produced in microgravity environments \cite{hofmann2015microgravity}.

The majority of metals in low earth orbit are diamagnetic metals, e.g. aluminum, due to their widespread use in aerospace\cite{opiela2009study}. These materials will see no appreciable reaction to a static magnetic field. They will however, react inductively to oscillating magnetic fields because they are electrical conductors. Conductive materials have a long history of positioning and heating from oscillating currents in electromagnetic levitators (EML) \cite{okress1952electromagnetic, wroughton1952technique, sagardia1977electromagnetic, spitans2017large}. These devices leverage a time-varying magnetic field, generated from a sinusoidal current in an electromagnet, that induces eddy currents in a conductive sample. These currents then heat the material due to ohmic losses and can position the sample to a local field minimum due to Faraday's law of induction \cite{griffiths2017introduction}. This phenomenon has even been demonstrated in low Earth orbit, first as part of the TEMPUS program on the space shuttle \cite{tempus2006containerless}\cite{egry2001containerless}, and more recently as part of the Materials-Science Laboratory Electromagnetic Levitator (MSL-EML) aboard the International Space Station (ISS) \cite{diefenbach2020experiment}\cite{matson2017use}.

EML designs are optimized around a static point where the sample can be positioned stably. Dynamic movement in 3D space is something that has been largely avoided in EML. But in order to recycle aluminum in microgravity, the samples (in both their solid and molten state) must be accurately positioned and moved through the system to facilitate processing. To that end, Pham et al. have shown that dexterous manipulation of conductive samples is possible using a rotating dipole field \cite{pham2021dexterous, dalton2022attracting, tabor2022adaptive}. Their model has been demonstrated to accurately position copper spheres in a simulated microgravity environment, and the force and torque model has been derived using dimensional analysis. However their method does not leverage the principal of superposition to increase field magnitudes and gradients by using coils harmoniously.

This paper intends to build upon the authors previous research that shows a force equation derived directly from the induced eddy current density can position a conductive sample in a simulated low-gravity environment \cite{stewart2025open}. Placing a semi-buoyant sphere in a water bath, open-loop control of aluminum subjected to time-varying magnetic fields was demonstrated through the creation of magnetic stable islands. These islands of local energy minima were generated using a four-coil array of electromagnets. The conductive samples, which are attracted to the local energy minimum due to their diamagnetic-like nature, were positioned in 3-dimensional space by moving these stable islands through manipulation of the magnetic fields. This paper extends that model by using positional feedback to apply appropriate forces for more accurate position control of conductive, non-magnetic samples, and explores how different choices in inverting the force equation lead to different overall performance.  

Throughout this work standard font, e.g. $a$ or $A$, represents scalars, bold font, e.g. $\mathbf{A}$, represents vectors, and the black-board font, e.g. $\mathbb{A}$, represents matrices. The identity matrix will be denoted $\mathbb{I}$ with the size implied by usage. The euclidean cross product will be written in its vector form as $\mathbf{a}\times\mathbf{b}$, or equivalently in its matrix-packed form as $\left[\mathbf{a}\right]_{\times} \mathbf{b}$. Functions of a variable, e.g. $f\{a\}$, will use curly brackets to identify the function inputs. Matrix element subscripts, e.g. $\mathbb{A}_{r,c}$ will follow the row-column convention. Time averaged quantities will be denoted with an overbar, e.g. $\overline{a}$. Imaginary numbers ($\sqrt{-1}$) will be denoted with the letter $i$.

%%%%%%%%%%%%%%%%%%%%%%%%%%%%%%%%%%%%%%%%%%

\section{Interaction Model} \label{sec:Theory}

\subsection{Force on an Induced Dipole} \label{subsec:Force on an Induced Dipole}
The potential energy $U$ of a magnetic dipole $\textbf{m}$ in a magnetic field $\textbf{B}$ is given by

\begin{equation}
    U = -\textbf{m} \cdot \textbf{B}.
    \label{eq:magnetic_potential_energy}
\end{equation}

\noindent By Faraday's law of induction, a changing magnetic field induces an electric field \cite{bauer2011university}, and a conductor in this electric field will see induced electrical current that creates a magnetic field to oppose the one that created it. From \cite{griffiths2017introduction} it is shown that the magnetic dipole can be found from the volume integral of the cross product of position $\textbf{r}$ in spherical coordinates and current density $\textbf{J} \unit({A/m^2})$

\begin{equation}
    \textbf{m} = \frac{1}{2} \int_V (\textbf{r} \times \textbf{J}) d\tau.
    \label{eq:moment_integral}
\end{equation}

\noindent For a sphere in a uniform oscillating field, Nagel \cite{nagel2017induced} found from the vector potential of the magnetic field that the induced current density in a sphere is described by the expression, in spherical coordinates,

\begin{equation}
    \mathbf{J}\{r,\theta\}= B_0 \left (\left[\frac{-3i\omega \sigma a}{2kaj'_1\{ka\} + 4j_1\{ka\}} \right] j_1\{kr\} \sin(\theta) \right)\hat{\phi}
    \label{eq:Nagel_inducedCurrentDensity}
\end{equation}

\noindent where $B_0$ is the magnitude of the magnetic field (\unit{T}), $\omega$ is the frequency of field oscillations (\unit{rad/sec}), $\sigma$ is the electrical conductivity of the object in the field (\unit{S/m}), $a$ is the sphere radius (\unit{m}), $j_1$ and $j'_1$ are the spherical bessel function of the first kind and its derivative, respectively, and $k$ is the complex propagation constant, which simplifies to 

\begin{equation}
    k = \sqrt{-i\omega\sigma\mu_0},
    \label{eq:wave_constant_k}
\end{equation}

\noindent at lower frequencies \cite{bidinosti2007sphere}\cite{nagel2017induced}. In \eqref{eq:wave_constant_k} $\mu_0$ represents the permeability of free space (\unit{$N/A^2$}). Combining \eqref{eq:moment_integral} and \eqref{eq:Nagel_inducedCurrentDensity} gives the following complex representation for the induced magnetic moment 

\begin{equation}
    \textbf{m}_c= \left (\frac{2\pi i \omega \sigma a}{k^4} \left ( a^2k^2 - 3 + 3ak \cot(ak) \right ) \right )\textbf{B} 
    \label{eq:complex_dipoleGain}
\end{equation}

\noindent which can be compactly represented as a complex dipole gain $g_c$ multiplied by the magnetic field vector

\begin{equation}
    \textbf{m}_c = g_c \textbf{B}.
    \label{eq:compact_complexMoment}
\end{equation}

It was shown in \cite{stewart2025open} that the complex valued dipole moment $\textbf{m}_c$ can be time averaged over one period, and the resulting average dipole moment $\bar{\textbf{m}}$ will scale linearly with the field value. This field value is itself a function of time, and can be expressed as 

\begin{equation}
    \textbf{B}\{t\} = \textbf{B}_a \sin(\omega t)
    \label{eq:time_dependent_field}
\end{equation}
\noindent where $\textbf{B}_a$ is the ampitude of the field in each direction. Expressing the dipole moment as a function of time yields 

\begin{equation}
    \textbf{m}\{t\} = \left\lVert g_c \right\rVert \textbf{B}_a \sin(\omega t+\gamma)
    \label{eq:time_dependent_moment}
\end{equation}

\noindent where here the term $\gamma$ represents the phase shift from the impressed sinusoidal field to the induced sinusoidal eddy current. The expected, or average, potential energy $\bar{U}$ seen in \eqref{eq:magnetic_potential_energy} can be expressed as 

\begin{equation}
   \bar{U} = -\left\lVert g_c \right\rVert\textbf{B}_a^T\textbf{B}_a  \frac{1}{T} \int_0^T \sin(\omega t + \gamma) \sin(\omega t) dt.
\end{equation}

\noindent Our convention has been to group the integral and magnitude of the complex gain seen into one term $\alpha$, as

\begin{equation}
    \alpha =\left\lVert g_c \right\rVert \frac{1}{T} \int_0^T \sin(\omega t + \gamma) \sin(\omega t)
    \label{eq:alpha_as_integral}
\end{equation}

\noindent which evaluates to

\begin{equation}
    \alpha = \left\lVert g_c \right\rVert \frac{\cos{\gamma}}{2}.
    \label{eq:alpha}
\end{equation}

 \noindent It is therefore appropriate to rewrite the magnetic potential energy equation \eqref{eq:magnetic_potential_energy} as being the average potential energy

\begin{equation}
    \bar{U} = -\alpha \textbf{B}_a^T \textbf{B}_a .
    \label{eq:induced_dipole_energy}
\end{equation}

As force is the negative gradient of potential energy

\begin{equation}
    \bar{\textbf{F}} = -\nabla \bar{U}
    \label{eq:general_force_from_energy}
\end{equation}

\noindent by the product rule the force on an induced magnetic dipole is therefore the product of the field and field gradient. By the zero-curl, zero-divergence criteria for a magnetic field, the 3x3 gradient matrix is symmetric and has zero trace, and can therefore be repackaged into a 5x1 vector $\textbf{G}$ where the full gradient is represented in just 5 terms \cite{petruska2015minimum}\cite{abbott2020magnetic}. The average force on an induced magnetic dipole on a magnetic field can therefore be written as 

\begin{equation}
    \bar{\textbf{F}} = 2\alpha [\textbf{B}]_f \textbf{G}
    \label{eq:force_equation_no_current}
\end{equation}

\noindent where the magnetic field vector has been repacked into a 3x5 matrix by the force packing operator, shown in the following expanded force equation 

\begin{equation}
    \begin{aligned}
        \bar{\mathbf{F}} = 2\alpha \begin{bmatrix} B_x & B_y & B_z&  0& 0 \\ 0& B_x &0 &B_y &B_z \\ -B_z& 0& B_x&     -B_z& B_y
            \end{bmatrix} \begin{bmatrix}
            \frac{\partial B_x}{\partial x} \\ \\ \frac{\partial B_x}{\partial y} \\ \\ \frac{\partial B_x}{\partial z} \\ \\ \frac{\partial B_y}{\partial y} \\ \\ \frac{\partial B_y}{\partial z}.
            \end{bmatrix}
    \label{eq:force}
    \end{aligned}
\end{equation}

\subsection{Dipole Approximation and the Superposition of Magnetic Sources} \label{subsec:Dipole Approximation}
The field generated from a magnetic source is calculated from the multipole expansion shown in \cite{griffiths2017introduction}, but higher order terms in the expansion drop off rapidly, and therefore if sufficiently far away from the source the magnetic field is accurately represented in the point dipole equation 

\begin{equation}
    \mathbf{B} = \frac{\mu_0}{4\pi \lVert \textbf{r} \rVert ^3} \left( 3 \hat{\textbf{r}} \hat{\textbf{r}}^T - \mathbb{I}_3 \right) \textbf{m}_s
    \label{eq:dipole_field}
\end{equation}

\noindent where $\textbf{r}$ is the distance between the dipole source $\textbf{m}_s$ and the point of interest. The gradient from a point dipole is found by the equation 

\begin{equation}
    \mathbf{G} = \frac{3 \mu_0}{4\pi \lVert \textbf{r} \rVert ^4} \left( \hat{\textbf{m}}_s\hat{\textbf{r}}^T + \hat{\textbf{r}} \hat{\textbf{m}}_s^T + \left (\hat{\textbf{r}}^T \hat{\textbf{m}}_s\right) \left( \mathbb{I}_3-5\hat{\textbf{r}} \hat{\textbf{r}}^T\right) \right).
    \label{eq:dipole_gradient}
\end{equation}

The field and gradient contribution from a dipole source obey the principle of superposition and can be added for multiple sources. It is therefore convenient to construct an actuation matrix $\mathbb{A}$ where each column contains the control normalized (value for one unit of control) field $\mathbb{B}$ and gradient $\mathbb{G}$ contribution at a point for each source, and therefore has eight (8) rows (three field and five independent gradient terms) and as many columns as there are field sources. For an electromagnetic coil system with $n$ sources, the actuation matrix can be multiplied by the coil current vector $\textbf{I}$ to give the field and gradient vectors at that point. 

\begin{equation}
    \begin{bmatrix}
        \textbf{B} \\
        \textbf{G}
    \end{bmatrix} = \begin{bmatrix}
        \mathbb{B}\\
        \mathbb{G}
    \end{bmatrix} \textbf{I} = \mathbb{A} \textbf{I}
    \label{eq:Actuation_to_fieldGradient}
\end{equation}

\noindent By the above equation, both the field and gradient values at a point in space $p$ can therefore be represented by the product of their normalized components of the actuation matrix and the current vector 
\begin{equation}
    \mathbf{B}\{p, \mathbf{I}\} = \mathbb{B}\{p\}\mathbf{I}
    \label{eq:field_actuation}
\end{equation}%    
\begin{equation}
     \mathbf{G}\{p,\mathbf{I}\} = \mathbb{G}\{p\}\mathbf{I},
     \label{eq:gradient_Actuation}
\end{equation}

\noindent where the explicit dependence on position $p$ is noted for completeness. Taking this simplified expression for the field and gradient and its relation to the coil currents that are the magnetic sources, the force equation \eqref{eq:force_equation_no_current} can be expressed as 
\begin{equation}
    \bar{\textbf{F}} = 2\alpha [\mathbb{B}\textbf{I}]_f \mathbb{G} \textbf{I}.
    \label{eq:force_equation_WITH_current}
\end{equation}

\section{Methods} \label{sec:Methods}

\subsection{Five-Coil Manipulation System} \label{subsec:5-coil system}
%%%%%%%%%%%%%%%%%%%%%%%%%%%%%%%%%%%%%%%%%%%%%%%%%%%
% %%%%%%%%%%%%%%%%%%%%%%%%%%%%%%%%%%%%%%%%%%%%%%%%%%%
\begin{figure}[ht!]
    \centering
    \includegraphics[width=9cm]{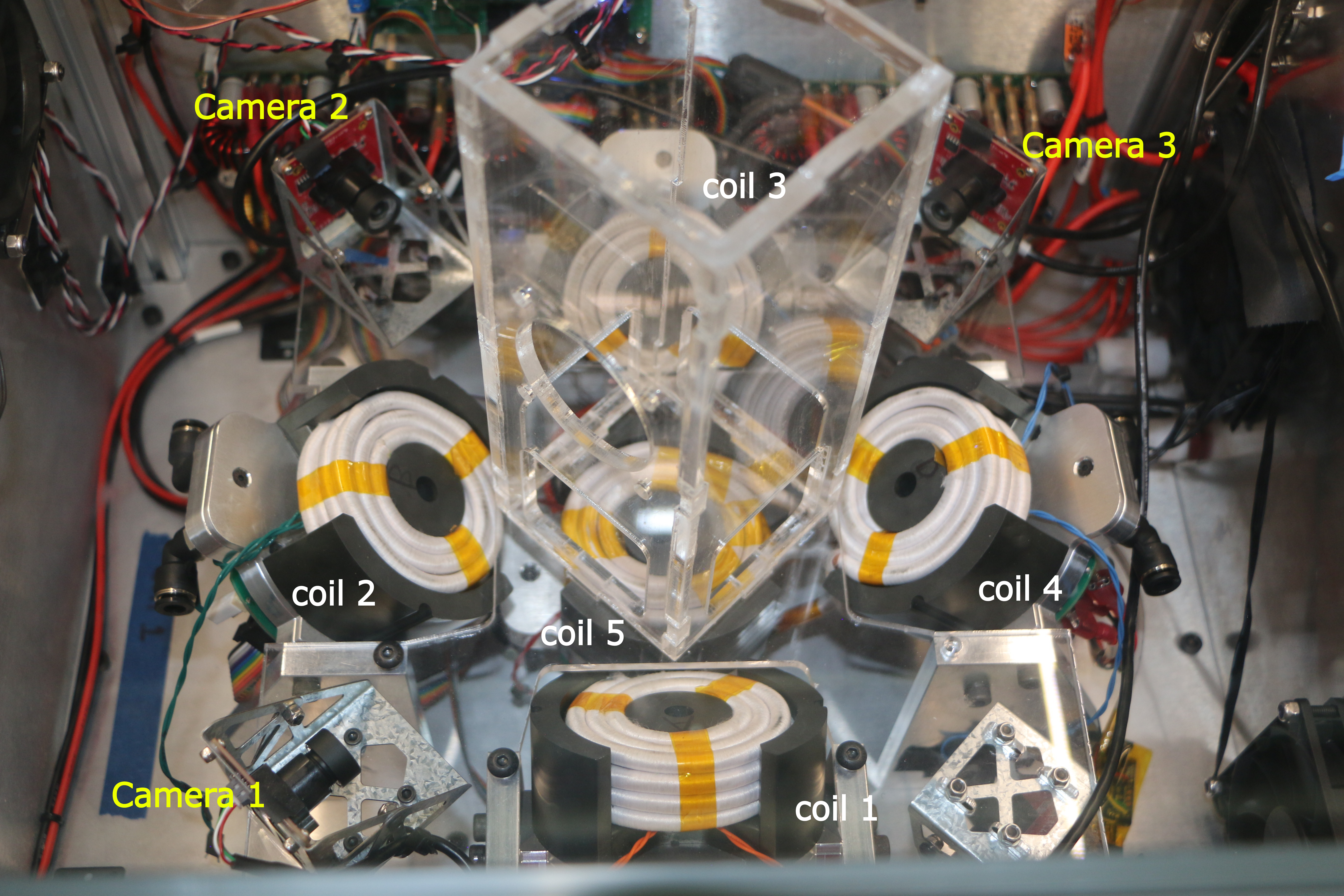}
    \caption{The five coil electromagnet array developed by CisLunar Industries for manipulation of conductive materials subject to oscillating magnetic fields. Coils 1-4 surround the workspace, while coil 5 is directly beneath the workspace. The magnetic workspace is surrounded by an acrylic enclosure for loading cartridges containing the samples. Three board-level cameras surround the working area, which are used for object detection in the feedback loop.}
    \label{fig:coilArray}
    % \vspace{-1em}
\end{figure}
% %%%%%%%%%%%%%%%%%%%%%%%%%%%%%%%%%%%%%%%%%%%%%%%%%%%
% %%%%%%%%%%%%%%%%%%%%%%%%%%%%%%%%%%%%%%%%%%%%%%%%%%%

In order to demonstrate the feasibility of the mathematical model described above, a five-coil electromagnetic array (MCEPC-STEER5x120-100, CisLunar Industries, Denver, CO., USA) was used to generate known field shapes surrounding a magnetic workspace (see Figure \ref{fig:coilArray}). The coils operate at approximately \unit[14.5]{kHz} and surround a calibrated magnetic workspace having approximate dimensions of \unit[5x5x5]{$cm^3$}. Each of the coils contains a soft-magnetic core and jacket to amplify the fields, and despite the strong coupling between sources each magnet is capable of independent amplitude and phase relative to the other four coils. Though the coils do have a ceiling of \unit[100]{A} on the allowable current that can be maintained in the coils, the independent control guarantees a certain level of flexibility when operating the system. 

Prior to operation, the system was calibrated by taking many different field measurements using a custom series of pickup coils (a similar method is described in detail in \cite{stewart2025open}). These field measurements were used to fit the coil system to a series of dipoles through non-linear least squares method \cite{petruska2017model}, where each magnetic source has an equivalent pose and strength to fit the measured data. The coefficients returned from the calibration process scale linearly with coil current, facilitating efficient dipole moment calculations for each magnetic source. The open-source software to calibrate a magnetic system is available to the community at \cite{Electromagnetic_Calibration_Repo}. The importance of the calibration process cannot be overstated, as it allows for the application of the relatively simple point-dipole model \cite{abbott2020magnetic} on a system that would otherwise require complex mathematical models not conducive to real-time control. For the experiments used in this paper, a calibration was used that had an $R^2$ value of 0.988, with most of the error found close to the fifth coil, which is located directly under (and close in proximity to) the bottom of the workspace. Fields and gradients are therefore confidently generated several centimeters above the bottom of the workspace.

%%%%%%%%%%%%%%%%%%%%%%%%%%%%%%%%%%%%%%%%%%%%%%%%%%%
% %%%%%%%%%%%%%%%%%%%%%%%%%%%%%%%%%%%%%%%%%%%%%%%%%%%
\begin{figure}[t!]
    \centering
    \includegraphics[width=9cm]{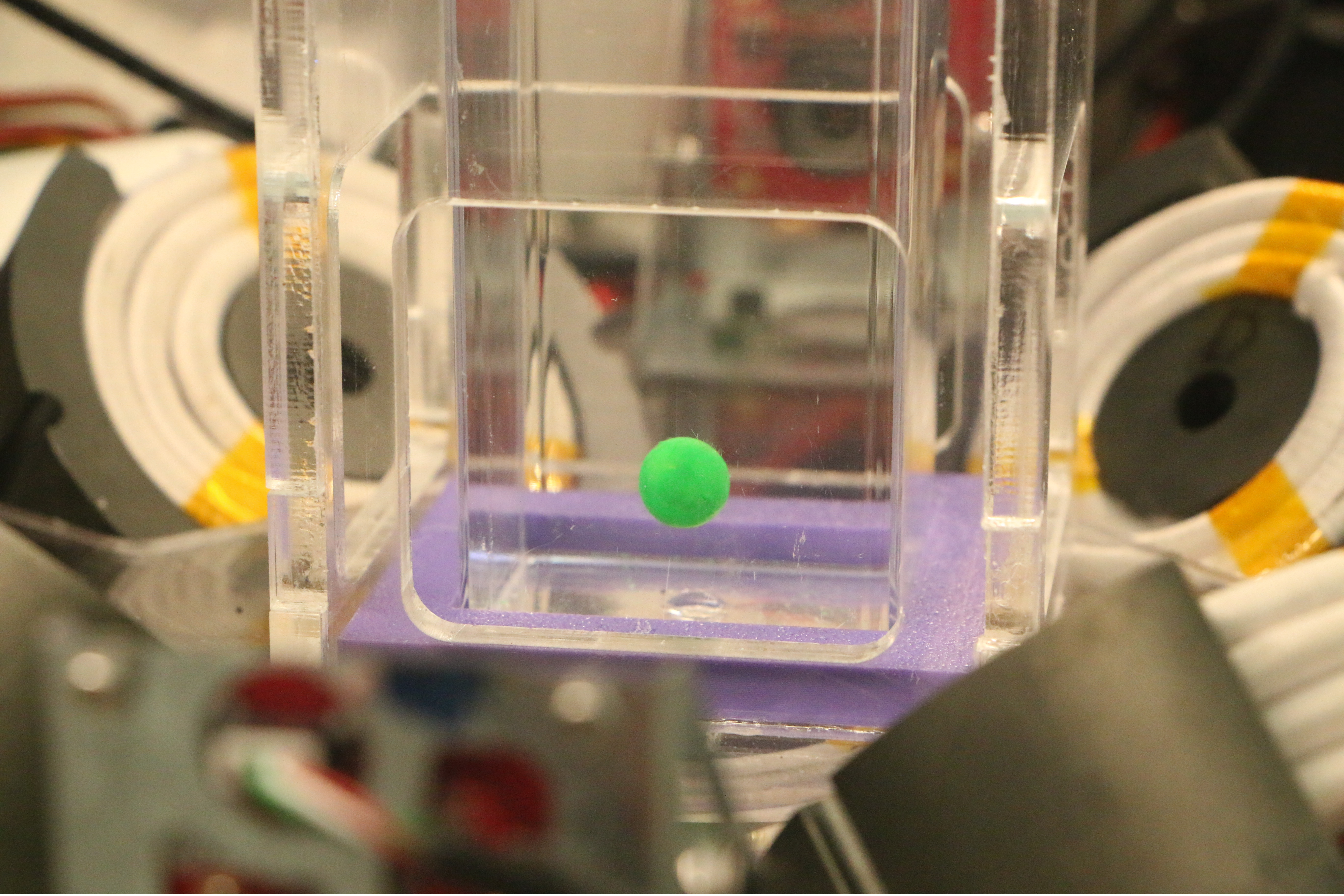}
    \caption{Simulated low-gravity environment to test coil-array in normal lab setting. Hollow aluminum sphere is partially filled with water and submerged in water bath, leveraging the buoyant force that pushes the sample to the surface and thereby allowing for levitation even under normal gravitational conditions. Typical effective sample weight is on the order of tens of micro-newtons.}
    \label{fig:glitterCaddy}
    % \vspace{-1em}
\end{figure}
% %%%%%%%%%%%%%%%%%%%%%%%%%%%%%%%%%%%%%%%%%%%%%%%%%%%
% %%%%%%%%%%%%%%%%%%%%%%%%%%%%%%%%%%%%%%%%%%%%%%%%%%%

\subsection{Sample Preparation} \label{subsec:sample preparation}

As mentioned, the five-coil array is limited in the amount of current that it can carry simply due to the saturation limit of the ferrite cores. As a result of this ceiling, and the inherently small value for the time-averaged dipole gain $\alpha$, the forces that the system is capable of generating are on the order of fractions of a millinewton. To test in a terrestrial laboratory it was necessary to simulate a low-gravity environment, which was achieved by filling a hollow aluminum sphere with enough water to make it nearly weightless when placed in a water bath, as can be seen in Figure \ref{fig:glitterCaddy}.

In our prior open-loop control work \cite{stewart2025open} it was critical to know the approximate weight of the ball in the water bath, as lifting forces in the reduced gravity environment were calculated \textit{prior to} running any trajectories. This is less critical for a closed-loop controller, which avoids the delicate tuning process altogether by simply including an integrator. One need only make sure that the effective weight of the sample makes it within the operable lifting range of the coils, which can be done by raising or lowering the temperature of the water, depending on the situation. It was found for this set of experiments, that slightly under-filling the hollow samples and then painting them until they sank produced the desired effective weight.

%%%%%%%%%%%%%%%%%%%%%%%%%%%%%%%%%%%%%%%%%%%%%%%%%%%
% %%%%%%%%%%%%%%%%%%%%%%%%%%%%%%%%%%%%%%%%%%%%%%%%%%%
\begin{figure*}[t!]
    % \centering
    \includegraphics[width=19cm]{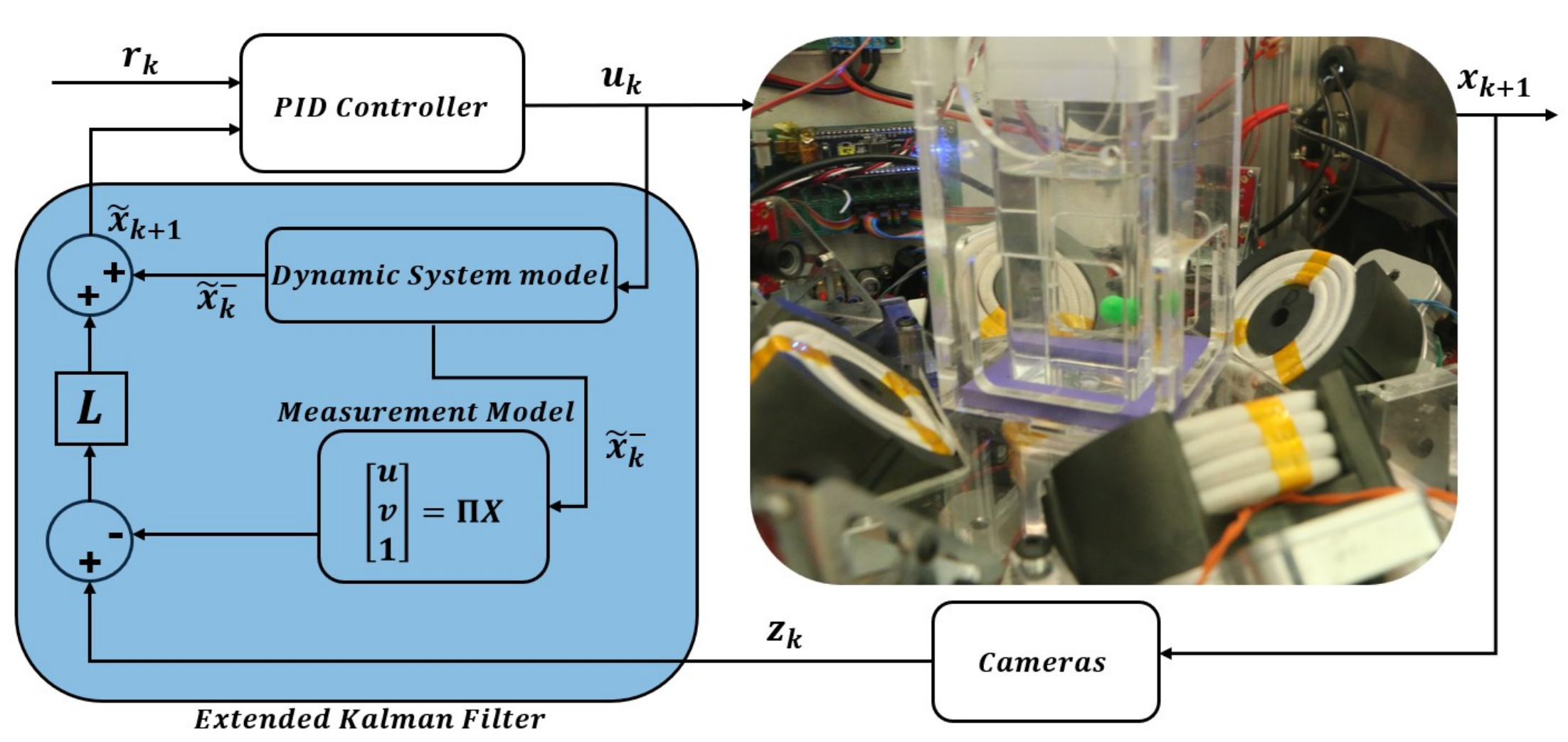}
    \caption{Control-loop employed in demonstrating closed-loop control of induced magnetic system. The observer, in the form of an Extended Kalman Filter, is shaded in blue, the output of which gives the state estimate of the sample during flight. The PID Controller takes as input the error in state and calculates a force value to apply that will reduce this state error. Feedback is provided by a set of three cameras, which provide the observer with an updated measurement of position.  }
    \label{fig:control_loop}

\end{figure*}
% %%%%%%%%%%%%%%%%%%%%%%%%%%%%%%%%%%%%%%%%%%%%%%%%%%%
% %%%%%%%%%%%%%%%%%%%%%%%%%%%%%%%%%%%%%%%%%%%%%%%%%%%

\subsection{Controller Design} \label{subsec:control loop}

Figure \ref{fig:control_loop} shows a high-level view of the controller used to generate the closed-loop position control for the induced magnetic system. The algorithm uses an Extended Kalman Filter (EKF) to estimate the state $\tilde{x}$ at each discrete update $k$ \cite{mohinder2001kalman}. A proportional-integral-derivative (PID) controller is used to convert the difference between state estimate and reference trajectory $r$ into control values $u$ at the current control update, where control is the desired magnetic force on the induced magnetic dipole. The magnetic force is applied by the plant, here modeled as a continuous second order system (more on the plant complexity below). The sample position is measured by cameras and the measurements $z$ are fed back into the EKF observer. The subsections below highlight the major components of the controller, which was implemented in C++ using the Eigen library \cite{eigenweb} for computation.

\subsubsection{Observer Dynamic System Model}
As mentioned above, an extended Kalman Filter \cite{welch1995introduction,simon2001kalman,mohinder2001kalman} is used to estimate the state of the system, where the governing differential equations are solved numerically using a Dormand-Prince 5th order solver \cite{dormand1980family}. The governing equation used to populate the state transition matrix is 

\begin{equation}
    \ddot{x} = \frac{1}{m}\left[F_{m}-F_{d}-F_{g}    \right]
    \label{eq:sum_forces}
\end{equation}

\noindent where $m$ is the mass of the object, $F_{d}$ is the fluid drag on the spherical sample \cite{gerhart2016munson}, estimated as 

\begin{equation}
    F_{drag} = 6 \pi \mu R  v
    \label{eq:stokes_drag}
\end{equation}

\noindent where $\mu$ in this context is the dynamic fluid viscosity, $a$ is once again the sphere radius, and $v$ is the relative fluid velocity. The term $F_{g}$ is the effective gravity after buoyant offset (typically orders of magnitude less than normal laboratory conditions). The magnetic force $F_{m}$ is simply the force on induced dipole, shown in \eqref{eq:force_equation_no_current}. For this version of the observer, the state vector contained six elements, three for position and three for velocity. The controller focused on position only, and neglected orientation control due to the symmetry of the spherical sample. 

To incorporate measurements into the observer three cameras are used to triangulate the position of samples as they move along trajectories. The Cameras were calibrated using a direct-linear transform \cite{DLT_CM}\cite{2023opencv}\cite{stereoCal} and the 3D affine transform \cite{besl1992method}, which produced a projection matrix $\Pi$ capable of transforming 3D position coordinates into pixel positions in the image plane. The spherical samples were detected using am image subtraction method \cite{kalsotra2022background} which combined blob detection and a moving window \cite{2023opencv}. The cameras used for image triangulation were all EV2U-SGR1-MMC1-C1 board level cameras from Innodisk USA Corporation, each fixed with CIL060 Low-Distortion lenses from Commonlands Optics. 

\subsubsection{Controller}
Estimated states were compared to reference trajectory to produce an error in position, and this position error was fed to a PID controller, as seen in Figure \ref{fig:control_loop}. Because the samples were not perfectly neutrally buoyant, it was necessary to include the integrator in the controller, as there was expected to be a steady-state force disturbance. 

Gains were calculated using robust pole placement method \cite{tits1996globally}. Each control update, the control law uses a discrete implementation by 

\begin{equation}
    u_k = K\left(x_{(des)k}- \tilde{x}_{k}   \right)
    \label{eq:control_law}
\end{equation}

\noindent where $K$ in this context is a matrix of controller gains. The updated control (force) values are then applied to the plant, producing a new state $x_{k+1}$ and camera measurement to inform the observer model, while also informing the dynamic model.

\subsubsection{Force Inversion Solver}
The force values that are controller outputs are turned into coil-currents through a Sequential Quadratic Programming (SQP) solver, which optimizes the problem
\begin{subequations}
    \begin{equation}
        \label{eq:ARGMIN_objFunction}
        \min ~ \left(\textbf{I} - \textbf{I}_r \right)^T \mathbb{W} \left(\textbf{I} - \textbf{I}_r \right)
    \end{equation}%
    \vspace{-2em}%
    \begin{align}%
        \label{eq:ARGMIN_c1}
         &~ \bar{\textbf{F}} - \textbf{F}_d = 0 \\
        \label{eq:ARGMIN_c2}
        &~ \bar{\textbf{F}} = 2\alpha [\mathbb{B}\textbf{I}]_f \mathbb{G} \textbf{I} \\
        \label{eq:ARGMIN_c3}
        &~ \textbf{I}_i \leq I_{max} \\
        \label{eq:ARGMIN_c4}
        &~ \textbf{I}_i \geq I_{min}
    \end{align}%
    \label{eq:ARGMIN}%
\end{subequations}%

\noindent In \eqref{eq:ARGMIN},  $\textbf{I}$ is the electrical current to be used in the next control update, while $\textbf{I}_r$ is a current reference value, $\mathbb{W}$ is a weighting matrix, $\bar{F}$ is the force value calculated from \eqref{eq:force_equation_WITH_current} and $F_d$ is the desired force output from the controller. As was mentioned above, the five coil system used for experimentation has hard limits on the maximum current that can be applied, and therefore in addition to the equality constraint that requires that the above equation satisfy the desired force, there is the requirement that current values do not exceed $\pm$ \unit[100]{A}.

% %%%%%%%%%%%%%%%%%%%%%%%%%%%%%%%%%%%%%%%%%%%%%%%%%%%
% %%%%%%%%%%%%%%%%%%%%%%%%%%%%%%%%%%%%%%%%%%%%%%%%%%%

\begin{figure}[t!]
    \centering
    \includegraphics[width=9cm]{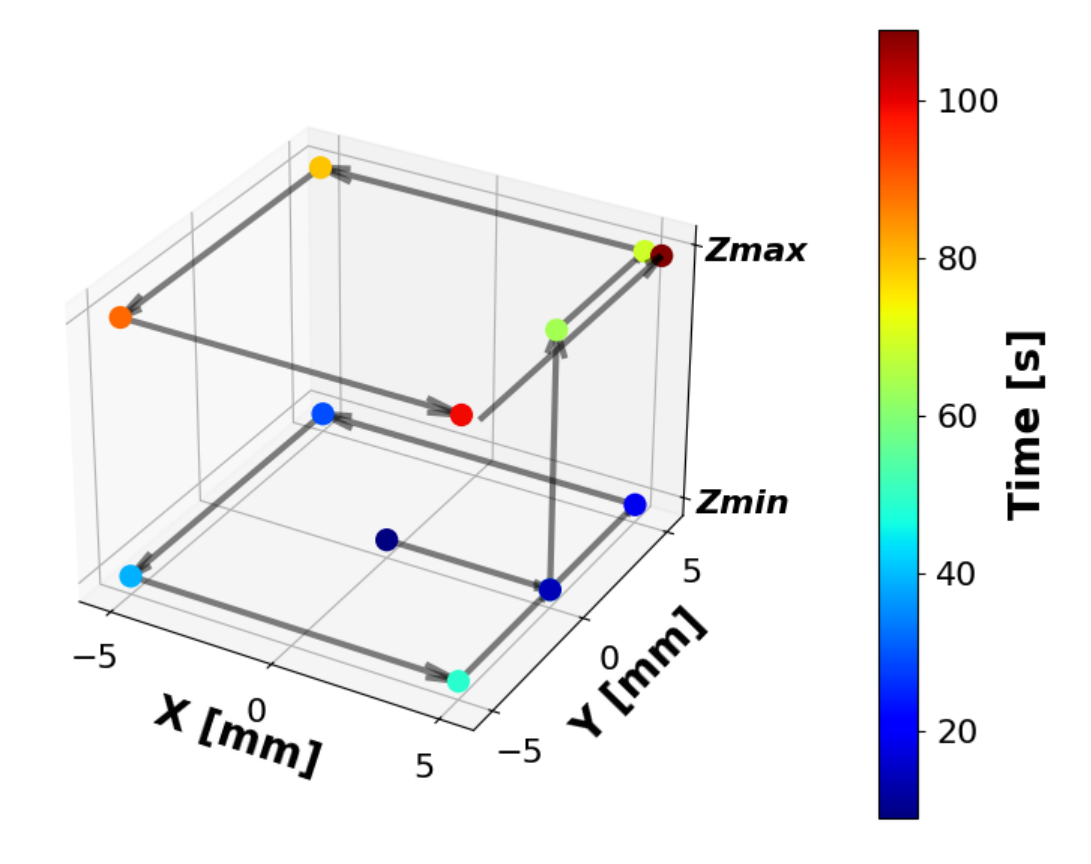}
    \caption{Planned trajectory in 3D space for the aluminum samples to travel during experimentation. After settling in the center of the workspace, samples will draw a \unit[10]{mm} square at the lower elevation, after which they will move \unit[10]{mm} in the positive z-direction and draw another \unit[10]{mm} square. Zmin/Zmax are \unit[20]{mm}/\unit[30]{mm} for the five-coil system, and \unit[52]{mm}/\unit[62]{mm} for the four-coil system. Total time to complete the cube is \unit[109]{seconds}. Offset in the final leg of the trajectory is not indicative of actual planned path, and is displayed in this manner so as to avoid overlapping paths.  } 
    \label{fig:planned_trajectory}
    \vspace{-1em}
\end{figure}

% %%%%%%%%%%%%%%%%%%%%%%%%%%%%%%%%%%%%%%%%%%%%%%%%%%%
% %%%%%%%%%%%%%%%%%%%%%%%%%%%%%%%%%%%%%%%%%%%%%%%%%%%

\subsection{Experimental Definition} \label{subsec:experimental definition}
Inductors cannot change current instantly \cite{bauer2011university} and therefore requesting large changes in current was something that was to be avoided when running the open-loop algorithm, as the time it took for the coils to reach steady-state current was enough time for the sample to become unstable and fall out of the workspace. In the open-loop trials it was therefore critical to keep the reference current $\textbf{I}_{r}$ as the solution found for the previous position in a trajectory, as this would minimize the chances of having \textit{coil-switching events}, defined as large, destabilizing changes in coil current. With the increased robustness afforded by the closed-loop controller comes increased freedom in how $\textbf{I}_{r}$ is defined, as closing the loop on control allows for the samples to recover after such a coil-switching event. This will allow for the solver to dictate controller behavior by defining alternative trajectories in 5D coil space based on preferred values for the reference current.

Three different experiments will be run, all varying the $\textbf{I}_{r}$ term described above. In varying this term, it is expected that the optimization procedure will direct system performance. In addition, an open-loop trajectory will be run with the 5-coil system in order to make a comparison between open-loop controller outlined in previous work \cite{stewart2025open}, and the closed-loop controller which is the focus of this study. 

In comparing control strategies, the same cube trajectory will be executed. The 3D trajectory, shown in Figure \ref{fig:planned_trajectory}, will draw a \unit[10]{mm} square at a height of \unit[20]{mm} before climbing to a \unit[30]{mm} elevation in the workspace and drawing another \unit[10]{mm} square, thereby completing the cube. Trajectories completed with the four-coil system are similar, though the elevations are slightly different (\unit[52]{mm}  for the lower square and \unit[62]{mm} for the upper square).  Prior to starting the trajectory, each sample is given \unit[10]{seconds} to settle at the center of the workspace from its arbitrary starting position at the beginning of the trial.  

\subsubsection{Minimizing Change in Current} \label{subsubsec: min_delta}
By setting the current reference to the previous current used, large changes in coil current can be avoided. This is the strategy that was employed in the open-loop trials, and will therefore serve as a baseline strategy when comparing controllers. Five trials will be run on each of the coil systems using this strategy.

\subsubsection{Minimizing Overall Current}
To maximize efficiency, current will be minimized during a set of trials by setting $\textbf{I}_{r}$ to the zero vector. In doing so, the solver will minimize the squared current as the sample completes the trajectory. It is anticipated that this may increase the number of switching events, and therefore comparing with a trajectory that sees very little (or no) switching events is ideal to get a statistical idea as to the positional sacrifice that is made in the name of efficiency. Five trials will be run that complete the same cube described for minimizing change in current, and statistical error for both sets of trials can then be compared.

%%%%%%%%%%%%%%%%%%%%%%%%%%%%%%%%%%%%%%%%%%%%%%%%%%%
%%%%%%%%%%%%%%%%%%%%%%%%%%%%%%%%%%%%%%%%%%%%%%%%%%%
\begin{figure*}[t!]
%\isPreprints{}{% This command is only used for ``preprints''.
% \begin{adjustwidth}{-\extralength}{0cm}
\centering
%\hfill
\subfloat[\centering]{\includegraphics[width=18.0cm]{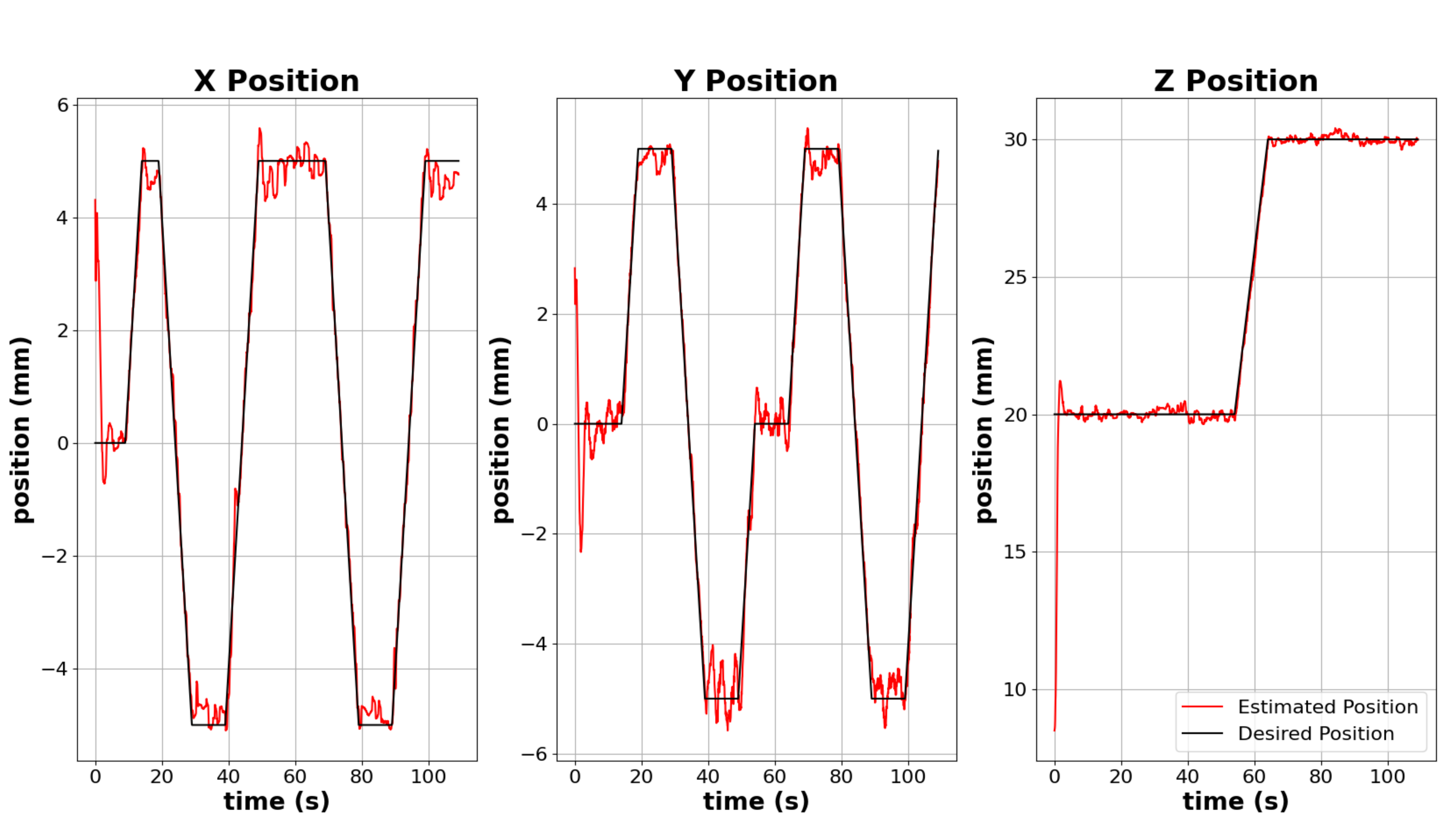}}\\
\subfloat[\centering]{\includegraphics[width=9.0cm]{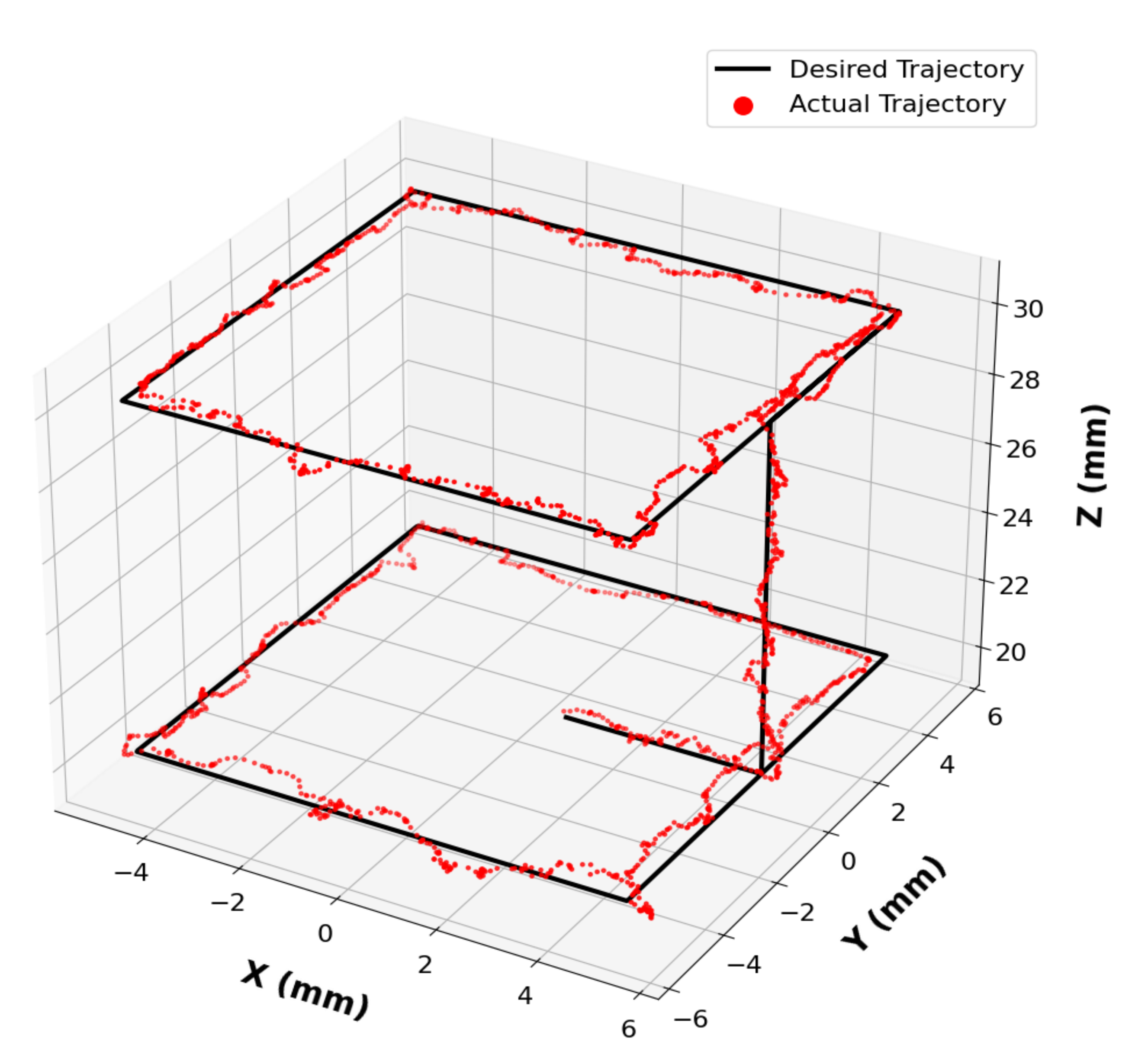}}
%\hfill
\subfloat[\centering]{\includegraphics[width=9.0cm]{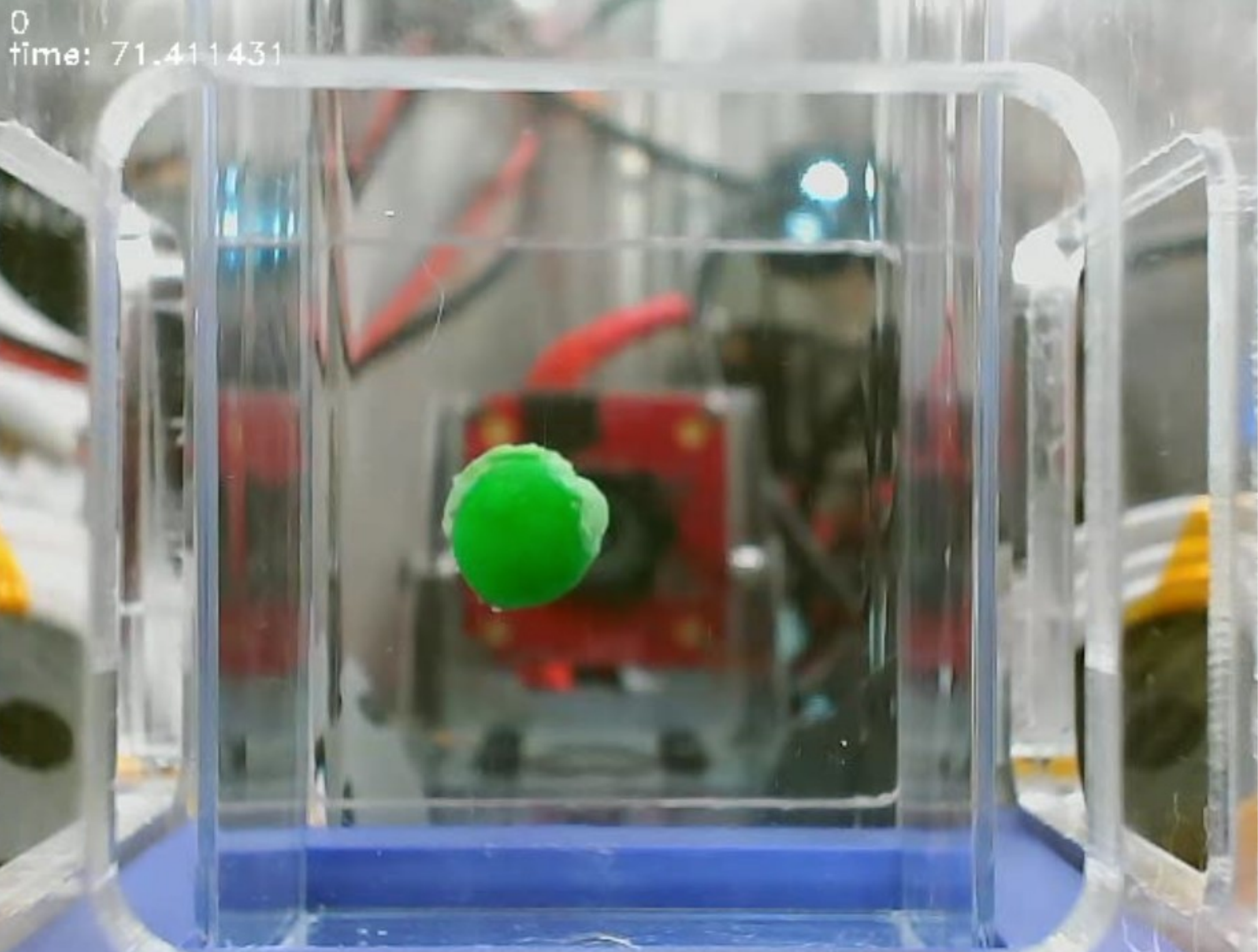}}
%\isPreprints{}{% This command is only used for ``preprints''.
% \end{adjustwidth}
\caption{(\textbf{a}) x,y, and z components for one of the trials that minimized change in current from one position to the next as a solver strategy. Desired position seen in black, with the estimated position returned by EKF shown in red. (\textbf{b})  Isometric view of the same trial, showing the full path of the ball as it completed the cube trajectory (first following the path of a square at the lower elevation of \unit[2]{cm}, and then moving to the higher elevation and following the path of the same square at a height of \unit[3]{cm}). (\textbf{c})  Still picture from the trial taken with one of the cameras used for triangulating position. Camera number and time seen in the top left corner.\label{fig:combinedResults}}
\end{figure*} 
%%%%%%%%%%%%%%%%%%%%%%%%%%%%%%%%%%%%%%%%%%%%%%%%%%%
%%%%%%%%%%%%%%%%%%%%%%%%%%%%%%%%%%%%%%%%%%%%%%%%%%%

%%%%%%%%%%%%%%%%%%%%%%%%%%%%%%%%%%%%%%%%%%%%%%%%%%
%%%%%%%%%%%%%%%%%%%%%%%%%%%%%%%%%%%%%%%%%%%%%%%%%%%

\begin{table}[!t]
%% increase table row spacing, adjust to taste
\renewcommand{\arraystretch}{1.5}
% if using array.sty, it might be a good idea to tweak the value of
% \extrarowheight as needed to properly center the text within the cells
\caption{RMSE and coil-switching while minimizing changes in coil current}
\label{tab:coilSwitchingRMSE}
\centering
% Some packages, such as MDW tools, offer better commands for making tables
% than the plain LaTeX2e tabular which is used here.
\begin{tabular}{|c||c||c|}
\hline
\textbf{Trial}	& \textbf{No. of Switches}	& \textbf{RMSE (mm)}\\
\hline
\hline
        1 & 0 & 0.448 \\
        \hline
        2 & 7 & 1.55 \\
        \hline
        3 & 0 & 0.522 \\
        \hline
        4 & 1 & 0.649 \\
        \hline
        5 & 3 & 0.853 \\
\hline
\end{tabular}
\end{table}
%%%%%%%%%%%%%%%%%%%%%%%%%%%%%%%%%%%%%%%%%%%%%%%%%%
%%%%%%%%%%%%%%%%%%%%%%%%%%%%%%%%%%%%%%%%%%%%%%%%%%%

\subsubsection{Open-Loop Reference Current Trajectory}  \label{subsubsec: ref_tracking}

For the closed-loop algorithms discussed in this paper so far, we have not concerned ourselves with finding stable solutions, as we are capable of applying any arbitrary force at a position in the workspace. This means that at any time during the trajectory we have hundreds of available current combinations that are capable of applying the same force on the sample. For the open-loop algorithm, the choices are much more limited. There are typically eight stable solutions available at any point in the workspace  for the previously used four-coil system \cite{stewart2025open}, though this increases to upwards of 20 or more solutions for the five-coil system.

Fundamentally, these stable solutions have a magnetic restoring force in all directions if the sample is perturbed from equilibrium. By generating an open-loop trajectory and using it as $\textbf{I}_{r}$ during a closed-loop trajectory, it is hypothesized that the currents returned from the solver will be close to these stable points and therefore be semi-stable. That is, the currents used during the closed-loop trajectory may not be technically stable but will likely be in a magnetic energy saddle and will benefit from additional magnetic restoring forces. This of course requires knowledge of the 5D coil trajectory before running the trials, as the time it takes to calculate stable solutions is not conducive to real-time control. Therefore if the strategy is successful in minimizing positional error, it comes at the cost of sacrificing some level of flexibility in planning. Five trials will be run using this strategy, determining whether the compromise is advantageous. 

\subsubsection{Comparison of Five-Coil and Four-Coil Systems}

The previous research into open-loop control used a four-coil system having a similar geometry as the five-coil system shown in Figure \ref{fig:coilArray}, with the omission of the coil underneath the workspace \cite{stewart2025open}. This paper will  explore all three closed-loop control strategies on both systems to determine if there is any significant difference in performance. Though both systems are capable of open-loop control, the open-loop experiments are highly susceptible to modeling errors and external disturbance, and therefore large differences in performance are equally likely to be due to environmental factors as they would be from actual performative factors.

\section{Results} \label{sec:Results}
%%%%%%%%%%%%%%%%%%%%%%%%%%%%%%%%%%%%%%%%%%%%%%%%%%
%%%%%%%%%%%%%%%%%%%%%%%%%%%%%%%%%%%%%%%%%%%%%%%%%%%
\begin{figure}[t!]
    \centering
    \includegraphics[width=8.5cm]{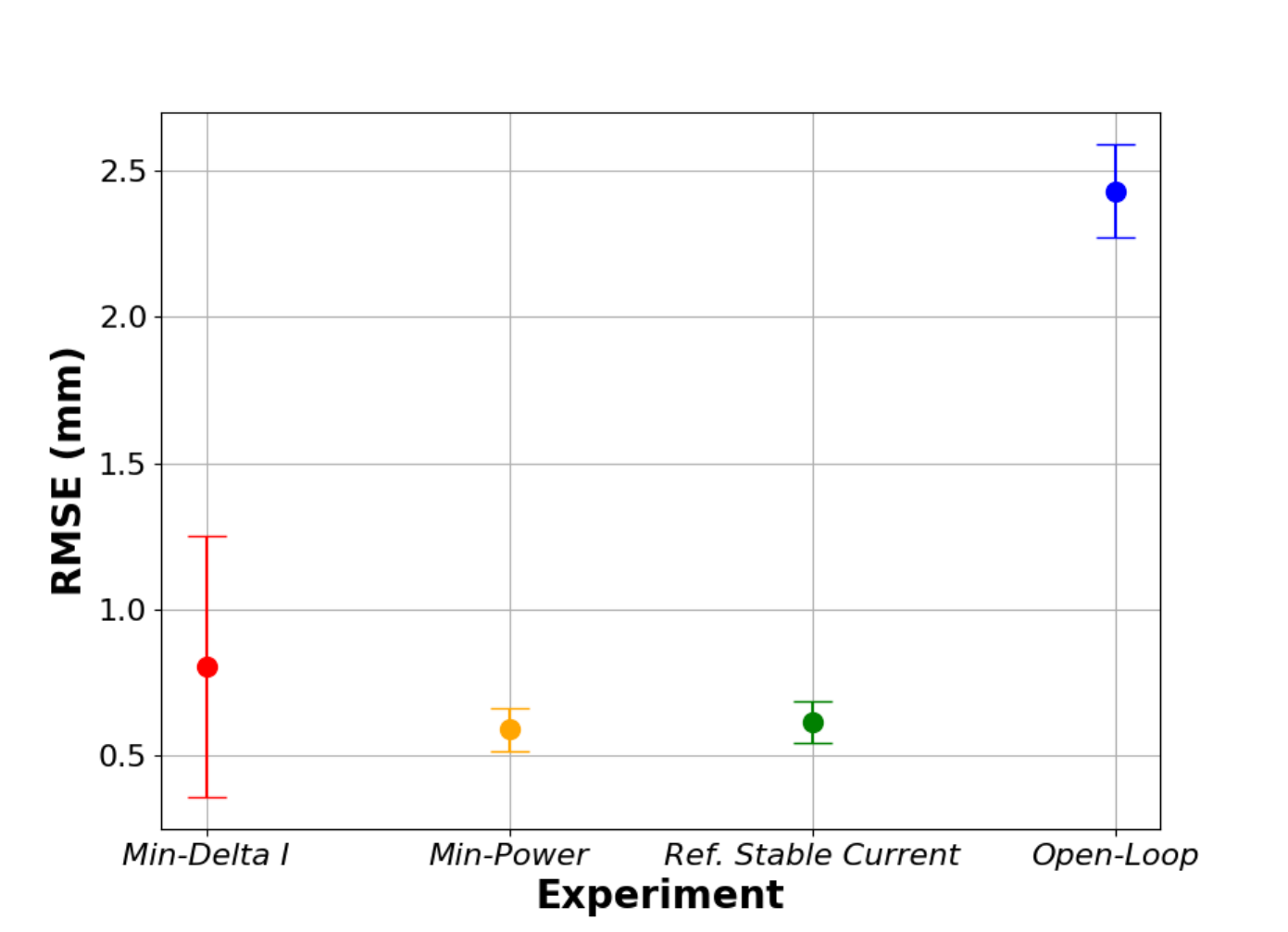}
    \caption{Root-mean squared error averaged over each trial, for each experimental setup. Error bar represents one standard deviation from the mean for all trials in the particular experiment.} 
    \label{fig:combinedRMSE}

\end{figure}
%%%%%%%%%%%%%%%%%%%%%%%%%%%%%%%%%%%%%%%%%%%%%%%%%%%
%%%%%%%%%%%%%%%%%%%%%%%%%%%%%%%%%%%%%%%%%%%%%%%%%%%

%%%%%%%%%%%%%%%%%%%%%%%%%%%%%%%%%%%%%%%%%%%%%%%%%%
%%%%%%%%%%%%%%%%%%%%%%%%%%%%%%%%%%%%%%%%%%%%%%%%%%%
\begin{figure}[t!]
    \centering
    \includegraphics[width=8.5cm]{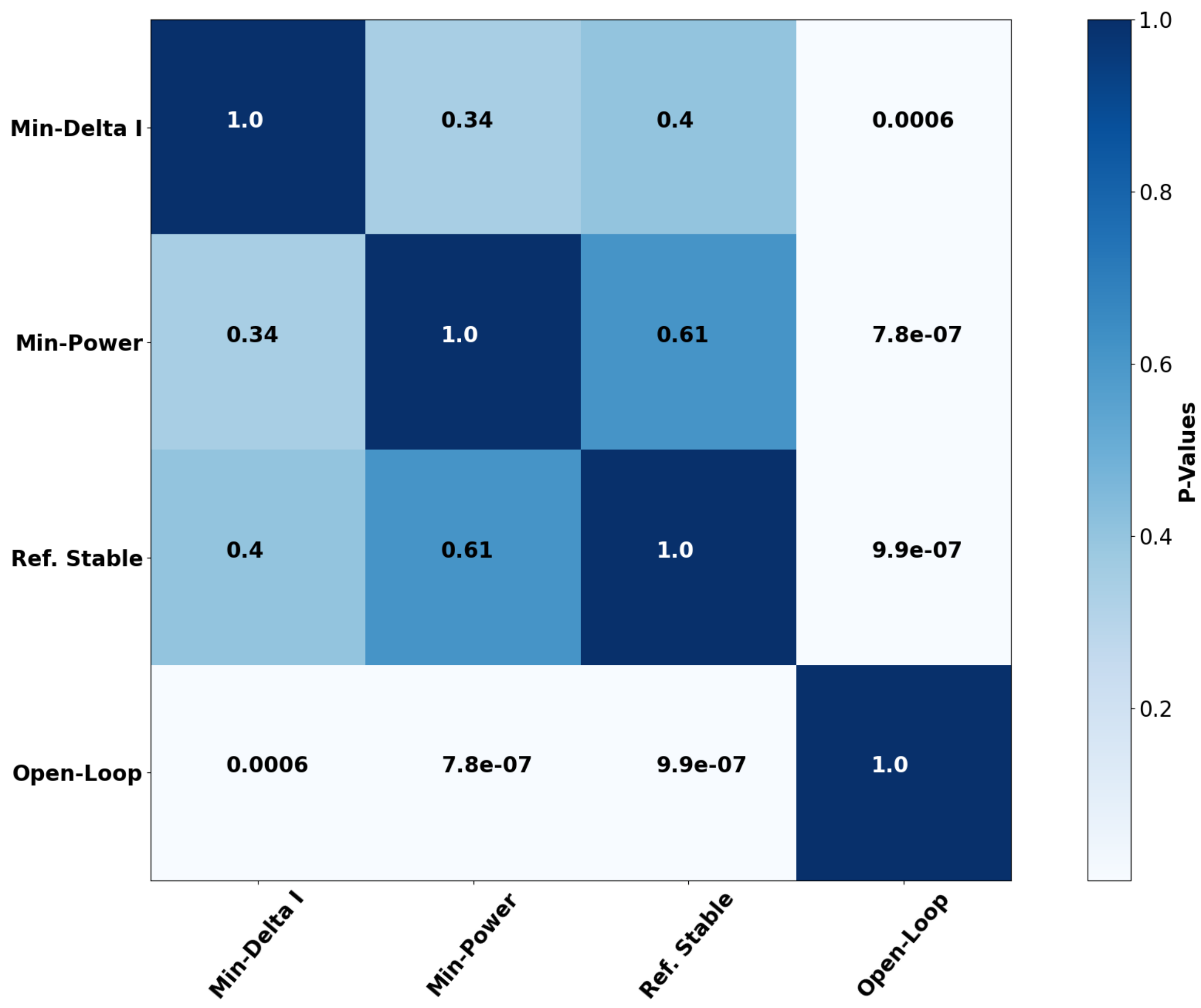}
    \caption{Calculated P-Values for RMSE for each control strategy. Values were calculated using a two-tailed t-test for each controller strategy while completing the cube trajectory with the 5-coil system. Using a statistical significance of 95$\%$, the plot demonstrates that only the open-loop trials show evidence that null-hypothesis can be rejected with confidence.  } 
    \label{fig:heatMap}
\end{figure}

%%%%%%%%%%%%%%%%%%%%%%%%%%%%%%%%%%%%%%%%%%%%%%%%%%
%%%%%%%%%%%%%%%%%%%%%%%%%%%%%%%%%%%%%%%%%%%%%%%%%%%

\subsection{Minimizing Change in Current} \label{subsec:minimizing change in current}

%%%%%%%%%%%%%%%%%%%%%%%%%%%%%%%%%%%%%%%%%%%%%%%%%%%
%%%%%%%%%%%%%%%%%%%%%%%%%%%%%%%%%%%%%%%%%%%%%%%%%%%
\begin{figure*}[t!]
    \centering
    \includegraphics[width=18cm]{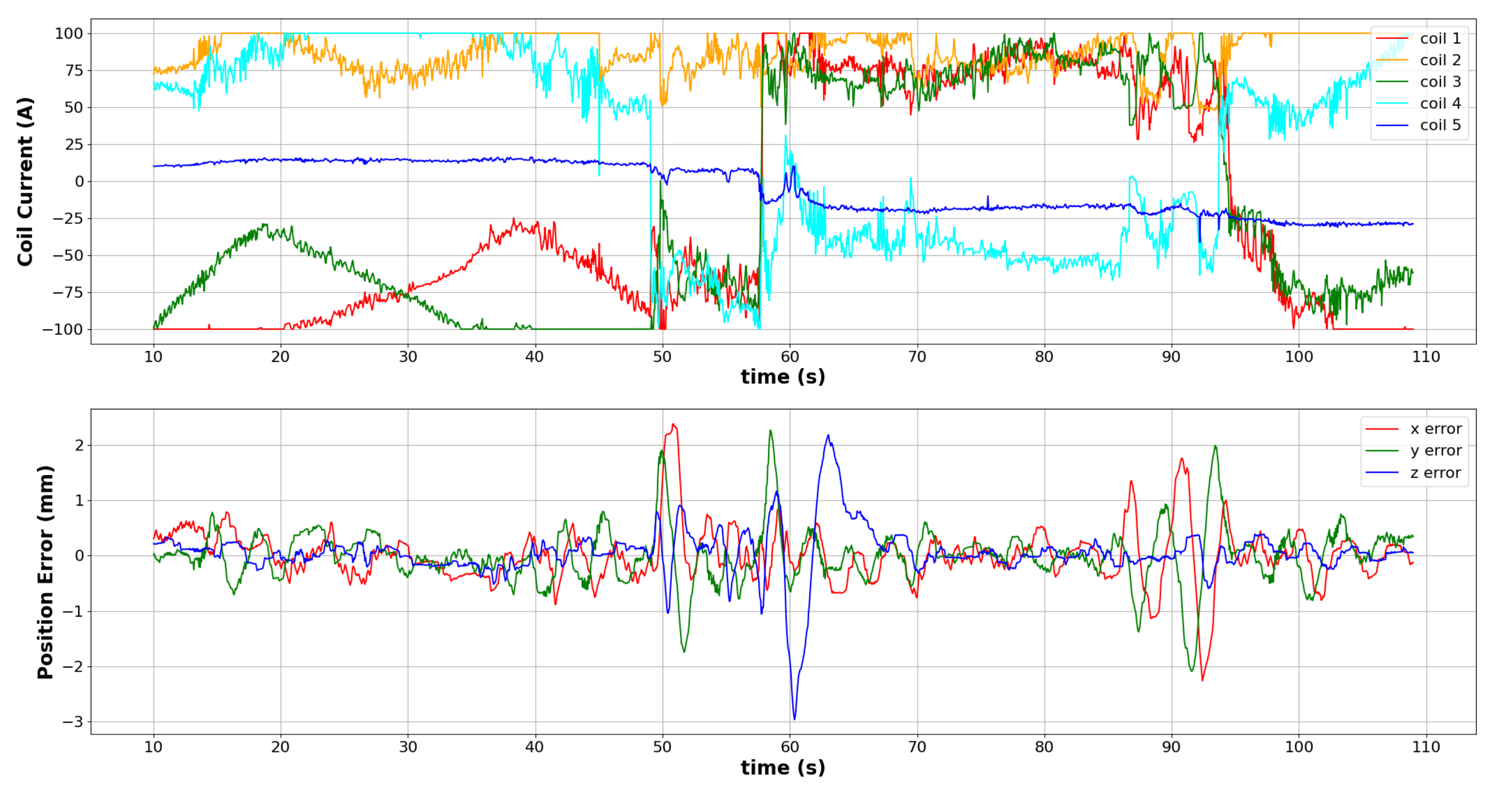}
    \caption{\textbf{Top}: Currents requested during trial 5 while minimizing change in current. Clear coil-switching events are seen at times 50, 58 and 95 seconds (roughly). \textbf{Bottom}: Error in position calculated while completing the trajectory using applied currents from top plot. Clear deviations from target trajectory are seen during and after coil-switching events, though sample recovers in all three instances.} 
    \label{fig:coilSwitch}
\end{figure*}
%%%%%%%%%%%%%%%%%%%%%%%%%%%%%%%%%%%%%%%%%%%%%%%%%%%
%%%%%%%%%%%%%%%%%%%%%%%%%%%%%%%%%%%%%%%%%%%%%%%%%%%

The results of a typical trial using the minimizing changes in current strategy (as described in \ref{subsubsec: min_delta}) can be seen in Figure \ref{fig:combinedResults}. This result is typical in that it demonstrates the same general behavior of the induced dipole that is seen in the other trials, and more importantly it does not have any coil-switching events, which will allow for a comparison to the behavior when large changes in coil current are present. As Figure \ref{fig:combinedResults} shows, the control strategy that minimizes changes in current from one control update to the next is able to accurately place the sample with respect to its desired position, with the average root-mean squared error (RMSE) over all five trials remaining below \unit[1]{mm}. This control strategy, though effective, does lead to coil-switching events, which increases the deviations from desired position. Table \ref{tab:coilSwitchingRMSE} shows how large changes in coil current can adversely affect positional accuracy, with the trials that experienced the most coil-switching events having the highest RMSE. It should be noted that samples in all trials were allowed to settle for the first ten seconds of the trajectory, as trials were started with the sample in any arbitrary position in the workspace which would have skewed the calculated RMSE. Therefore results for all trials consider times only after the first ten seconds of the trajectory.

Using this controller as a baseline, each control strategy's ability to track position can be compared. The RMSE was therefore calculated as the sample completed its trajectory for each of the five trials, for each of the four experimental setups. Figure \ref{fig:combinedRMSE} shows the average RMSE for each controller strategy, with the standard deviation plotted as error bars at each data point, while Figure \ref{fig:heatMap} shows p-values comparing the likelihood of statistically significant difference in experimental groups (using a significance level of $\alpha = 0.05$). As both of these plots demonstrate, the various closed-loop control strategies have all shown significant improvement when compared to open-loop strategies, though there is no definitive evidence that any closed-loop strategy tracks position better than the others.

The large variation seen while employing the controller that minimizes change in current is likely the result of the coil switching, as seen in Table \ref{tab:coilSwitchingRMSE}. These large changes in coil current are simply a function of what the solver requests to achieve a desired force, though the reason for this discontinuous request is multifaceted. The most common reason for requesting discontinuous current vectors is because the coils are operating near their current limits and run out of options that would allow effective minimization of the change from previous control updates. Another common reason to request large changes in coil currents are driven by object detection errors causing instantaneous changes in position seen by the observer. This positional uncertainty creates large swings in force requested by the controller which the solver accommodates with large changes in requested current. 

Regardless the cause, it should be noted that these coil-switching events were previously unrecoverable during the open-loop experiments, causing the ball to fall out of the workspace. For the present controllers, it is evident that these large changes in coil current are recoverable, as the sample will regain its position and continue on the trajectory after the event. This can be seen in Figure \ref{fig:coilSwitch} which clearly shows coil-switching events at roughly \unit[50, 58, and 95]{seconds}, each demonstrating an associated oscillation around its desired position. In all cases though, the sample stabilizes and is able to continue its trajectory.

\subsection{Minimizing Overall Current} \label{subsec:minimizing overall current}

As mentioned above, it was anticipated that by maximizing efficiency there would be a trade off in positional accuracy. That is, by constantly minimizing current to achieve the required restoring forces there would be many large changes in coil currents that would cause the sample to oscillate or even fall out of the workspace. This surprisingly was not the case at all, as there were more coil-switches while minimizing power (with an average of 5.8 switches over the five trials, compared with an average of 2.2 switching events for the baseline controller), but the positional accuracy did not suffer from these switching events. This is supported by Figure \ref{fig:heatMap}, which shows that there is no evidence that the RMSE while minimizing power is any different than when minimizing for change in current between controller updates. 

%%%%%%%%%%%%%%%%%%%%%%%%%%%%%%%%%%%%%%%%%%%%%%%%%%
%%%%%%%%%%%%%%%%%%%%%%%%%%%%%%%%%%%%%%%%%%%%%%%%%%%
\begin{figure}[t!]
    \centering
    \includegraphics[width=8.5cm]{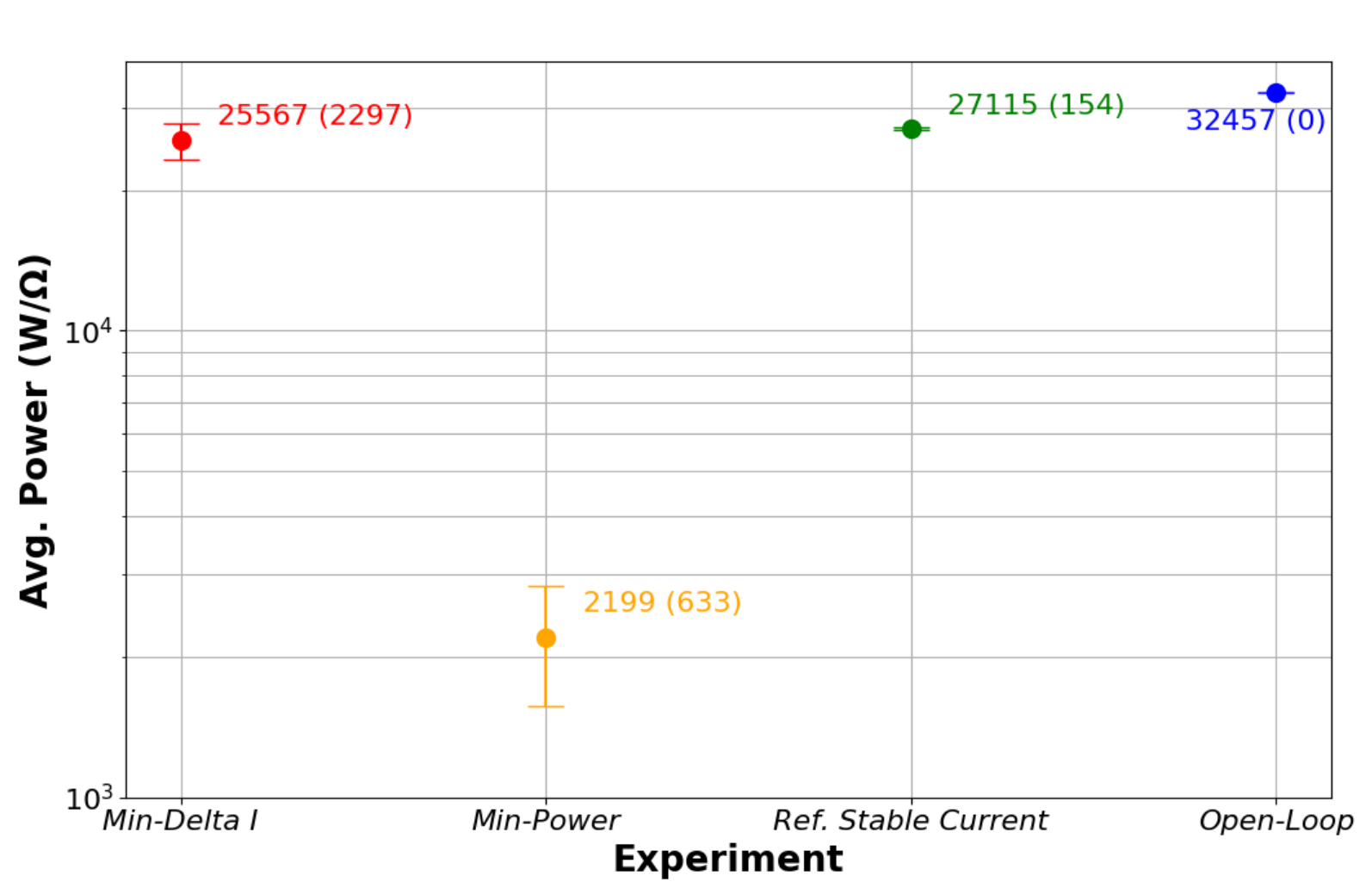}
    \caption{Average power consumption (normalized by coil resistance) over all trials for each test condition. Numbers next to data points are the calculated average over all five trials, with the term in parenthesis representing the standard deviation for all trials.} 
    \label{fig:powerComparison}
\end{figure}
%%%%%%%%%%%%%%%%%%%%%%%%%%%%%%%%%%%%%%%%%%%%%%%%%%%
%%%%%%%%%%%%%%%%%%%%%%%%%%%%%%%%%%%%%%%%%%%%%%%%%%%

%%%%%%%%%%%%%%%%%%%%%%%%%%%%%%%%%%%%%%%%%%%%%%%%%%
%%%%%%%%%%%%%%%%%%%%%%%%%%%%%%%%%%%%%%%%%%%%%%%%%%%

\begin{table}[!t]
%% increase table row spacing, adjust to taste
\renewcommand{\arraystretch}{1.5}
% if using array.sty, it might be a good idea to tweak the value of
% \extrarowheight as needed to properly center the text within the cells
\caption{RMSE and coil-switching while minimizing overall current draw in the coils}
\label{tab:coilSwitching_MIN_POWER}
\centering
% Some packages, such as MDW tools, offer better commands for making tables
% than the plain LaTeX2e tabular which is used here.
\begin{tabular}{|c||c||c|}
\hline
\textbf{Trial}	& \textbf{No. of Switches}	& \textbf{RMSE (mm)}\\
\hline
\hline
      1 & 5 & 0.601 \\
      \hline
        2 & 6 & 0.620 \\
        \hline
        3 & 9 & 0.689 \\
        \hline
        4 & 6 & 0.539 \\
        \hline
        5 & 3 & 0.500 \\
\hline
\end{tabular}
\end{table}
%%%%%%%%%%%%%%%%%%%%%%%%%%%%%%%%%%%%%%%%%%%%%%%%%%
%%%%%%%%%%%%%%%%%%%%%%%%%%%%%%%%%%%%%%%%%%%%%%%%%%%

Table \ref{tab:coilSwitching_MIN_POWER} shows how the control strategy of minimizing overall current was able to maintain a favorable RMSE value despite there being more coil-switching events. The reason for this seemingly contradictory result is that though there were changes in current, the changes were not as great as with other test set-ups because the magnitude of current was always much smaller. That is, the transient times associated with coil-switches were minimized because the current amplitudes are simply smaller, and therefore have a shorter distance to travel in coil-current space. 

It should be noted that with these trials identifying just what constituted a coil-switching event became much more ambiguous. When the solver requested an entirely new current vector for other experiments, there is a clear shifting from positive-to-negative (or negative-to-positive) for one or more coils, but when minimizing current the clarity of this shift is harder to see, as it is difficult to tell if this is a coil-switch or a natural change in coil-trajectory. Further complexity is found when the solver may briefly request a swing in current, but then naturally returns to the smaller current in one or two control updates (as its goal is always to find the value that is closest to zero), and because the actual current in the coils changes slower than these anomalous updates it acts as a low-pass filter and is largely unaffected by these temporary swings. Regardless of the vague nature of these coil-switches, it is clear that minimizing overall current in the coils has the surprising effect of minimizing error in position for this set of trials, indicating that it may be an effective strategy for controlling position in and of itself.

Figure \ref{fig:powerComparison} demonstrates average power consumption for all trials and compares over each experimental setup. Average  normalized power for each trial was calculated from 

\begin{equation}
    \bar{P} = \frac{1}{n} \sum_{k=1}^{n} \textbf{I}_k^T \textbf{I}_k
    \label{eq:averagePower}
\end{equation}

\noindent where $\textbf{I}_k$ is the $kth$ current vector in the trajectory, $n$ is the total number of control updates during the trial, and $\bar{P}$ is the average power (normalized by resistance in the coils). Using this formula for each trial, an average power over all trials can be calculated for each experimental condition.

As the plot in Figure \ref{fig:powerComparison} demonstrates, the strategy of minimizing power shows an order of magnitude improvement on efficiency when compared to all other testing conditions. From these set of experiments it is therefore clear that by minimizing the total current in the coils, one would benefit from not only a more efficient machine concerning power consumption, but would also see a strong control authority over samples simply due to the minimized amplitudes and subsequent minimized transient times.

\subsection{Reference Current Tracking} \label{subsec:reference current tracking}

The hypothesis outlined in Section \ref{subsubsec: ref_tracking} states that using an open-loop current trajectory as a reference input to equation \eqref{eq:ARGMIN} would add an additional passive restoring force to the sample as it moved through its 3D trajectory. According to Figure \ref{fig:combinedRMSE} this strategy does not appear to have made any appreciable improvements when compared to the baseline strategy of minimizing change in current, which is supported with a p-value of 0.61 shown in Figure \ref{fig:heatMap} suggesting that any differences in controller strategies could be explained by noise. The implication of this result is that though the open-loop stability can maintain a sample's position, its ability to passively maintain position is not comparable to the active ability of a closed-loop controller. From this perspective, it is therefore hard to justify the time spent calculating a trajectory offline prior to running the trials. 

%%%%%%%%%%%%%%%%%%%%%%%%%%%%%%%%%%%%%%%%%%%%%%%%%%
%%%%%%%%%%%%%%%%%%%%%%%%%%%%%%%%%%%%%%%%%%%%%%%%%%%
\begin{figure*}[t!]
    \centering
    \includegraphics[width=18cm]{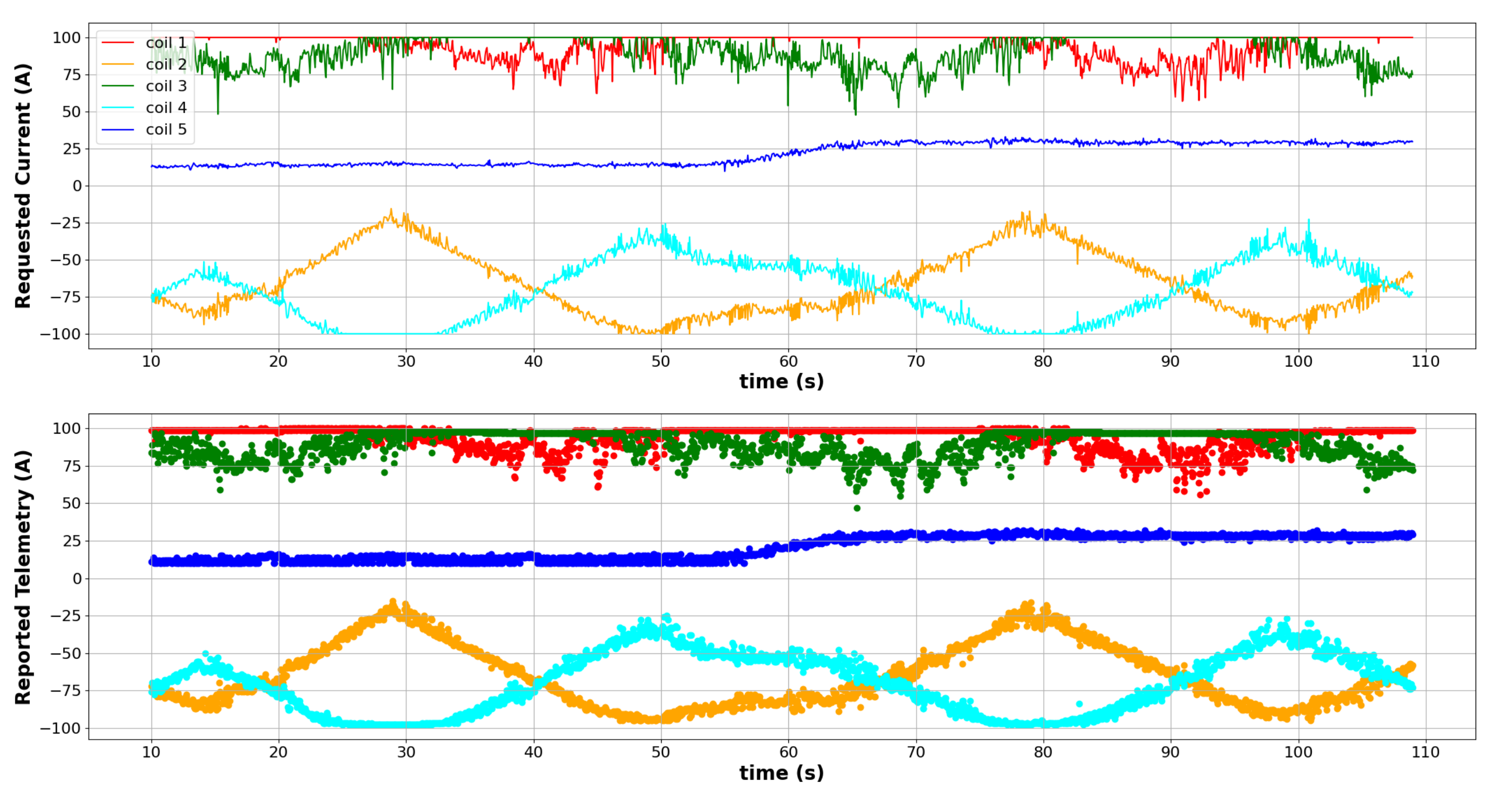}
    \caption{\textbf{Top} Current trajectory used as a reference when completing a cube trajectory for the "Reference Current Tracking" trials. Trajectory is an open-loop current trajectory that stably supports the weight of a \unit[56]{$\mu$ N}  sample as it moves through the workspace. \textbf{Bottom} Actual currents reported back from the 5-coil system as the ball completed its cube trajectory. Notice that coil-switching events are absent, as was the case for all trials in this particular experiment. }
    \label{fig:ref_tracking_telemetry}
\end{figure*}
%%%%%%%%%%%%%%%%%%%%%%%%%%%%%%%%%%%%%%%%%%%%%%%%%%%
%%%%%%%%%%%%%%%%%%%%%%%%%%%%%%%%%%%%%%%%%%%%%%%%%%%

One asset the reference tracking has is that it avoids coil-switching events entirely, as the solver always has a continuous ground-truth current trajectory to return to when solving for appropriate currents. Figure \ref{fig:ref_tracking_telemetry} shows the open-loop trajectory used as $\textbf{I}_r$ when attempting to leverage semi-stable positions during experimentation, and compares these values with the reported telemetry (actual measured coil currents).  As the figure demonstrates, there was very little deviation from the reference current as the ball completed its 3D path through the workspace. Note the lack of coil-switching events, which was the case for all reference tracking trials. Due to the predictable currents used in this experiment, it may prove an effective strategy in mitigating unwanted behavior during any process that has little variability (on an assembly line, for instance), though its improvement on position control beyond the baseline strategy of minimizing change in current is not strongly supported by this research. In addition, since the open-loop current trajectory assumes a perfect knowledge of the magnetic field state and restricts available current choices to this idealized trajectory, it makes the strategy less robust to calibration errors.

\subsection{Open and Closed-Loop Comparisons} \label{subsec:OL_CL comparison}

\begin{figure*}[t!]
    \centering
    \includegraphics[width=18cm]{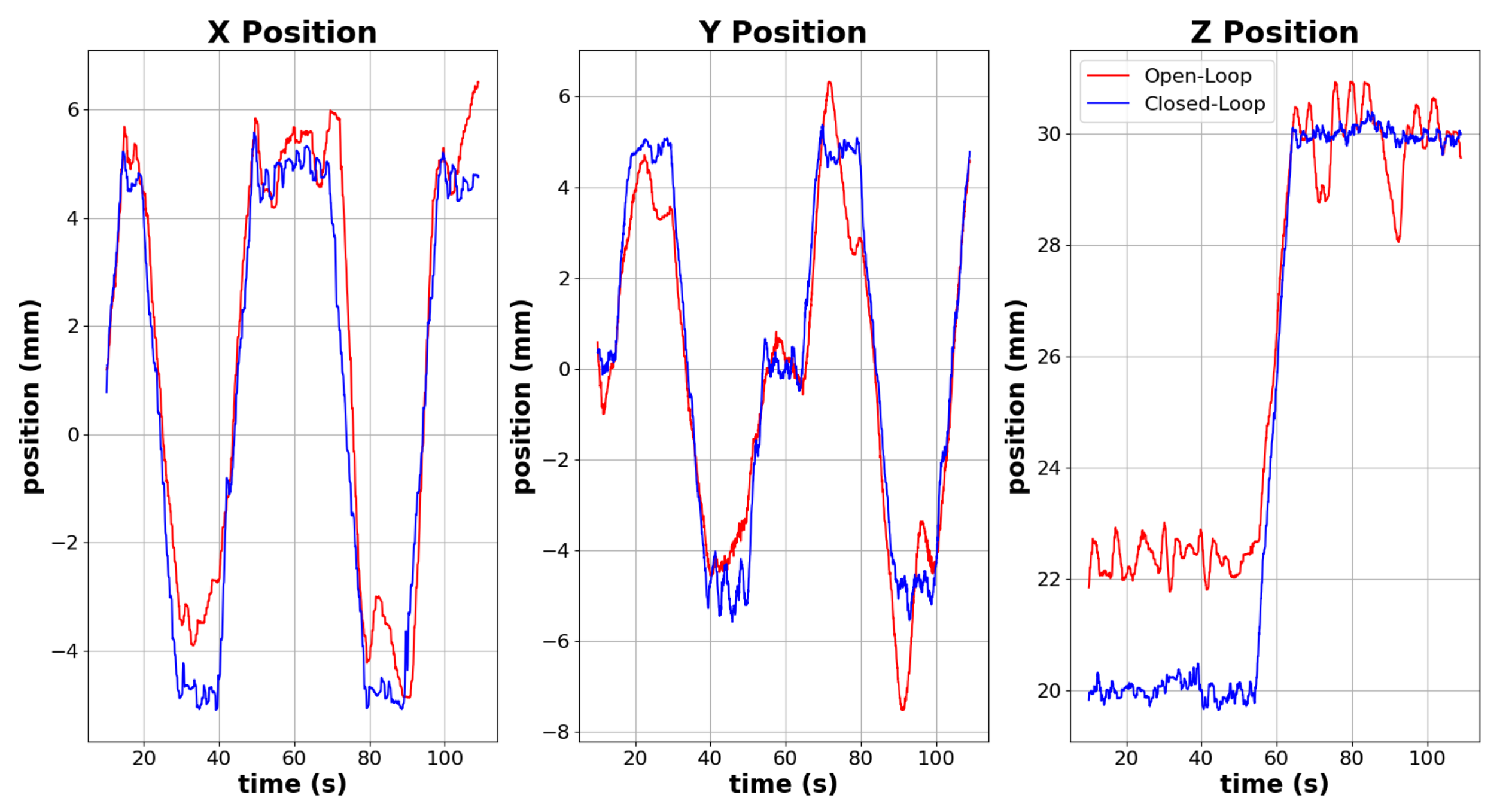}
    \caption{Comparison of  typical open-loop controller results (red) and  typical closed-loop controller that minimizes change in current (blue). RMSE for open and closed-loop controllers for these particular trials was \unit[2.3]{mm} and \unit[0.4]{mm}, respectively.}
    \label{fig:OL_CL_comparison}
    % \vspace{-1em}
\end{figure*}

As Figure \ref{fig:combinedRMSE} shows, there is significant improvement in position tracking when comparing the closed-loop controllers to the open-loop controller (supported by the statistical analysis shown in Figure \ref{fig:heatMap}, where the open-loop controller's p-value is several orders of magnitude below the 0.05 threshold that supports rejection of the null-hypothesis). Figure \ref{fig:OL_CL_comparison} shows both an open-loop trajectory and a closed-loop trajectory side-by-side. From the plot it is clear that a closed-loop strategy minimizes overshoot and undershoot seen in the open-loop case, and steady-state offset is eliminated entirely. It is interesting though how well the open-loop controller tracks x and y positions, as it is clear from this plot that most of the open-loop RMSE can be attributed to the steady-state offset in the z-position during the first minute of the trajectory. 

The open-loop controller used for this five-coil system does also show qualitative improvement on the open-loop controller seen in \cite{stewart2025open}, which had significant over-shoot and offset at times. It is believed that the addition of the fifth coil under the workspace improves control authority, as it has been shown that increased number of magnetic sources has an overall improvement on a system's ability to generate desired forces \cite{petruska2015minimum}. It would therefore be an interesting future direction to explore the efficacy of open-loop position tracking for system with various number of electromagnetic coils, though it is outside of the scope of this paper. 

\subsection{Four-Coil and Five-Coil Comparison} \label{subsec:4_coil_comparison}

%%%%%%%%%%%%%%%%%%%%%%%%%%%%%%%%%%%%%%%%%%%%%%%%%%
%%%%%%%%%%%%%%%%%%%%%%%%%%%%%%%%%%%%%%%%%%%%%%%%%%%
%%%%%%%%%%%%%%%%%%%%%%%%%%%%%%%%%%%%%%%%%%%%%%%%%%%
%%%%%%%%%%%%%%%%%%%%%%%%%%%%%%%%%%%%%%%%%%%%%%%%%%%

%%%%%%%%%%%%%%%%%%%%%%%%%%%%%%%%%%%%%%%%%%%%%%%%%%
%%%%%%%%%%%%%%%%%%%%%%%%%%%%%%%%%%%%%%%%%%%%%%%%%%%
\begin{figure}[t!]
    \centering
    \includegraphics[width=8.5cm]{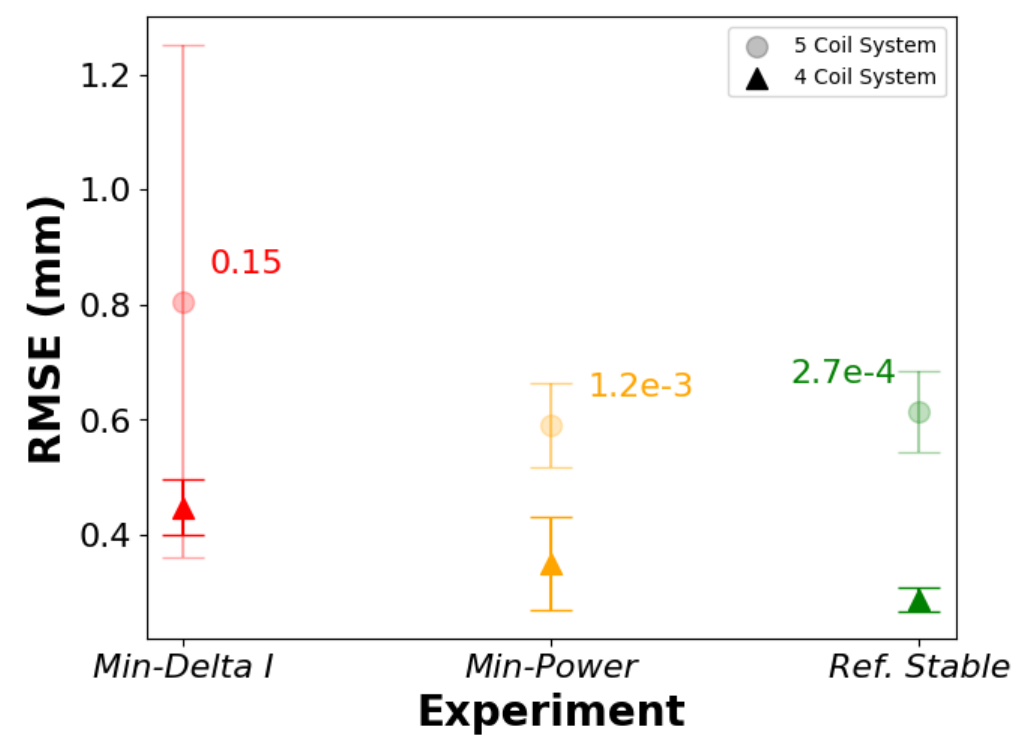}
    \caption{Root-mean squared error for both the 4-coil and 5-coil systems for each closed-loop controller strategy. P-values written next to each system comparison were generated using a two-sample t-test using a significance level of $\alpha = 0.05$. } 
    \label{fig:systemComparisons}
    % \vspace{1em}
\end{figure}
%%%%%%%%%%%%%%%%%%%%%%%%%%%%%%%%%%%%%%%%%%%%%%%%%%%
%%%%%%%%%%%%%%%%%%%%%%%%%%%%%%%%%%%%%%%%%%%%%%%%%%%

Moving from a four-coil system to a five-coil system broadens the capabilities of the system, allowing for more complex field shapes and an increased degree of freedom. It was suspected that this additional coil would also either increase positional tracking ability, or have no effect at all on position error for the closed-loop controllers. As Figure \ref{fig:systemComparisons} shows, this is not the case, as the four-coil system shows decreased RMSE when compared to the five-coil system at the heart of this analysis. This surprising result becomes less so when considering the fundamental differences in the systems. The four-coil system uses two Black Fly cameras with Navitar lenses, whereas the five-coil system uses three board level cameras with much lower resolution. Considering all average RMSE values remain sub-millimeter, the improved result could therefore simply be due to improved fidelity in object-detection. 

Another likely source of variation is the calibration of each coil system, with the four-coil having a higher $R^2$ when recreating the electromagnets as a series of dipoles through calibration. Mathematically, the field shapes predicted by the four-coil system are going to be closer to the fields and gradients realized by the system. In addition, the physical position of the fifth coil relative to the other four coils make it highly susceptible to cross-coupling. This at times causes the fifth coil to be pulled to unintended current magnitudes, generating fields that may not be advantageous to course correction. 

The p-values shown in Figure \ref{fig:systemComparisons} suggest that the controller that minimizes power and the controller that uses a stable trajectory as a reference current both show statistically significant differences in systems, while the controller that minimizes change in current between control updates does not show any statistical difference in the mean RMSE. One possible reason for this overlap in the base-controller case is the large standard deviation seen in the five-coil system, which had several trials showing a large RMSE due to coil switching events. Though these switching events seem to occur by chance, they can be influenced by environmental factors such as lighting and sample weight relative to water temperature. The presence of large error for this set of trials suggests that conditions may have been more favorable during the four-coil trials, though more testing is warranted to determine if these systems are truly indistinguishable for this controller strategy. 

It should be noted that though there may be differences in the system performance, the five-coil system remains much more flexible in its ability to adapt to variations in lab conditions. From experience with both systems it has been found that the general trend for the five-coil system to adapt to fluctuations in lab temperature or changes in sample weight, for instance, is less likely to result in a failed experiment. The four-coil system, on the other hand, is much more like threading a needle, as it typically operates close to the threshold of its ability. Despite that, as is shown in Figure \ref{fig:systemComparisons}, when this needle is threaded, the system performs well.

\section{Conclusion} \label{sec:Conclusion}

In demonstrating closed-loop control of an induced magnetic dipole subject to oscillating magnetic fields in  the workspace of a five-coil electromagnetic array, several strategies for inverting the force equation were explored and their effectiveness compared. Using an observer based controller that estimates the state of a semi-buoyant aluminum sphere using positional feedback from cameras, the force equation was inverted to determine the appropriate coil currents needed to generate the required forces. This study showed that strategies that use an open-loop current reference trajectory had no appreciable effect on the controller's ability to track position, though it did eliminate coil-switching events. The strategy that attempted to minimize power throughout the trajectory showed an order of magnitude improvement on efficiency, and surprisingly did not sacrifice position control for this gain in efficiency. As was expected, all closed-loop strategies showed improved control authority when compared to an open-loop trajectory. Minimal differences in performance can be seen when comparing the four-coil system with  a similar five-coil system, though these differences are likely the result of improved machine vision, electromagnetic calibration and uncontrolled mutual inductance between coils.

\section*{Acknowledgment}

The authors would like to thank Eugene Hamzezadeh for his support in generating the controls software.

\ifCLASSOPTIONcaptionsoff
  \newpage
\fi



\bibliographystyle{IEEEtran}
\bibliography{IEEEabrv,references.bib}
\end{document}